\documentclass[11pt, a4paper, logo, copyright, nonumbering]{style}
\usepackage[authoryear, sort&compress, round]{natbib}
\usepackage{graphicx}
\usepackage{subcaption}
\usepackage{xspace}
\usepackage{xcolor}
\usepackage{adjustbox}
\usepackage{array}
\usepackage{float}
\usepackage{geometry}
\usepackage{dblfloatfix}
\usepackage{enumitem}
\usepackage{setspace}
\usepackage{footmisc}
\usepackage{booktabs}
\usepackage{multirow}
\usepackage{longtable}
\usepackage{tabularx}
\usepackage{xltabular}
\usepackage{threeparttable}
\usepackage{siunitx}

\usepackage{amsmath}
\usepackage{amssymb}
\usepackage{amsfonts}
\usepackage{mathtools}
\usepackage{bm}
\usepackage{dsfont}

\usepackage{subcaption}
\usepackage{tikz}
\usepackage{pgfplots}
\pgfplotsset{compat=1.18}
\usepackage{algorithm}
\usepackage{algpseudocode}
\usepackage{wrapfig}
\usepackage{listings}

\usepackage{cleveref}
\usepackage{hyperref}
\hypersetup{colorlinks=true}

\usepackage{todonotes}
\usepackage{fontawesome5}
\usepackage[most]{tcolorbox}
\usepackage{CJKutf8}
\usepackage{lipsum}

\makeatletter
\def\@BTrule[#1]{%
  \ifx\longtable\undefined
    \let\@BTswitch\@BTnormal
  \else\ifx\hline\LT@hline
    \nobreak
    \let\@BTswitch\@BLTrule
  \else
     \let\@BTswitch\@BTnormal
  \fi\fi
  \global\@thisrulewidth=#1\relax
  \ifnum\@thisruleclass=\tw@\vskip\@aboverulesep\else
  \ifnum\@lastruleclass=\z@\vskip\@aboverulesep\else
  \ifnum\@lastruleclass=\@ne\vskip\doublerulesep\fi\fi\fi
  \@BTswitch}
\makeatother

\addto\extrasenglish{
}

 {\begin{list}{}%
         {\setlength{\leftmargin}{#1}}%
         \item[]%
 }
 {\end{list}}

\reportnumber{001}
\title{\centering Reasoning Path Divergence: A New Metric and Curation Strategy to Unlock LLM Diverse Thinking
}
\date{}

\author[*]{
        Feng Ju$^{1,2}$, 
        Zeyu Qin$^1$, 
        Rui Min$^1$, 
        Zhitao He$^1$, 
        Lingpeng Kong$^3$, 
        Yi R. (May) Fung$^1$
\\
\small $^1$The Hong Kong University of Science and Technology \\~~~
\small $^2$University of Science and Technology of China \\~~~
 \small       $^3$The University of Hong Kong \\~~~
}

\vspace{-10pt}

\correspondingauthor{$~$Equal Contribution.}

\renewcommand{\phi}{\varphi}

\renewcommand{\epsilon}{\varepsilon}
\renewcommand{\imath}{\mathrm{i}}

\newlength{\restsubwidth}
\newlength{\restsubheight}
\newlength{\restsubmoreheight}
\newcommand{\rest}[2]{%
        \settowidth{\restsubwidth}{\ensuremath{#2}}
        \settoheight{\restsubheight}{\ensuremath{{}_{#2}}}
        \ensuremath{{#1\hskip 0.5pt}_{\vrule\kern2pt\parbox[b][%
        4pt][b]{\the\restsubwidth}{%
                        \ensuremath{{}_{#2}}}}}
        }
\begin{abstract}
While Test-Time Scaling (TTS) has proven effective in improving the reasoning ability of large language models (LLMs), low diversity in model outputs often becomes a bottleneck; this is partly caused by the common \textit{one problem, one solution} (1P1S) training practice, which provides a single canonical answer and can push models toward a narrow set of reasoning paths. This homogenization not only limits sampling effectiveness but also restricts the exploration space for subsequent Reinforcement Learning (RL) stages. To address this, we propose a \textit{one problem, multiple solutions} (1PNS) training paradigm that exposes the model to a variety of valid reasoning trajectories and thus increases inference diversity. A core challenge for 1PNS is reliably measuring semantic differences between multi-step chains of thought, so we introduce Reasoning Path Divergence (RPD), a step-level metric that aligns and scores Long Chain-of-Thought solutions to capture differences in intermediate reasoning. Using RPD, we curate maximally diverse solution sets per problem and fine-tune Qwen3-4B-Base. Experiments show that RPD-selected training yields more varied outputs and higher pass@k, with an average +2.80\% gain in pass@16 over a strong 1P1S baseline and a +4.99\% gain on AIME24, demonstrating that 1PNS further amplifies the effectiveness of TTS.

\vspace{15pt}
\textbf{Code}:  \url{https://github.com/fengjujf/Reasoning-Path-Divergence} \\
\textbf{Correspondence}: Yi R. (May) Fung 

\end{abstract}

\begin{document}
\begin{CJK*}{UTF8}{gbsn}

\maketitle

\section{Introduction}

Large Language Models (LLMs) \citep{llm_achiam2023gpt,llm_chowdhery2023palm,llm_touvron2023llama} have made significant progress on tasks requiring complex reasoning that were previously challenging for automated systems, including competition-level mathematics and theoretical physics \citep{wang2025letsreasonformallynaturalformal,huang2025adactrladaptivecontrollablereasoning}. Chain-of-Thought (CoT) prompting \citep{cot_wei2022chain,cot_nye2021show} plays an important role in that progress by eliciting step-by-step reasoning from language models. Built on CoT, Test-Time Scaling (TTS) methods, which have been widely adopted in recent research, further achieve substantial improvements on complex reasoning tasks, by generating multiple reasoning trajectories at inference time and selecting among them through techniques such as Best-of-N sampling \citep{BoN1_brown,BoN2_song} and self-consistency \citep{MV_wang,wang2025calmunleashingcrosslingualselfaligning}. However, the effectiveness of TTS methods critically depends on the diversity of generated reasoning paths \citep{CE_rethinking_DCO,CE_weight_ensembling,diversity_aware_PO,inference_aware_finetune}. If the model's reasoning paths show minimal variation, the gains from additional sampling remain limited.

This diversity bottleneck is partly a consequence of common training practices for reasoning: datasets typically pair each problem with a single solution, which, by habitually exposing the model to one pathway, effectively teaches it to converge on a single ``correct'' way of reasoning rather than to explore the space of valid alternatives. Consequently, models tend to adopt that canonical trajectory and seldom produce alternative reasoning paths. Such "mode collapse" in the SFT phase is particularly detrimental when the model serves as the initial policy for RL. If the starting policy is overfit to a single, homogenous solution path, the agent's exploration during RL is severely handicapped, often confining it to narrow, suboptimal local optima \citep{GEM, sft_foundation_for_rl}. Thus, establishing a diverse solution space in SFT is not merely an isolated improvement but a prerequisite for robust RL. Several works have sought to counteract this by modifying objectives or introducing diversity-aware losses \citep{GEM,CE_rethinking_DCO,diversity_aware_PO}, yet it is still unclear how the diversity in the training examples maps onto the diversity of output at test time. That uncertainty motivates our core question:

\begin{center}
    \textit{Can a one problem, multiple solutions training paradigm effectively mitigate output homogenization and improve TTS performance?}
\end{center}

In this work, we explore a pragmatic approach to address this diversity bottleneck: training models on datasets where each problem is paired with multiple distinct solutions. To build such datasets, we first need to solve a key challenge: how to reliably measure semantic diversity between complex reasoning paths. Common approaches, such as computing cosine similarity on embeddings  \citep{sentencebert} of the entire solution text, fail for Long Chain-of-Thought solutions because they conflate high-level strategic differences with low-level computational details and narrative style. To tackle this issue, we introduce Reasoning Path Divergence (RPD), a new diversity metric that uses large language models to summarize solutions into their core logical steps and then applies an asymmetric matching process to measure semantic overlap. This approach allows RPD to tell true strategic novelty from superficial variations, forming a basis for systematically curating diverse data.

Using this metric, we chose the OpenThought3 dataset \citep{openthoughtsguha2025} for our experiments. It contains 53,125 math problems, each with 16 long chain-of-thought answers. These features make the dataset a well-suited testbed for our experiments on diversity-driven data selection.

Our main contributions in this work are:
\begin{itemize}[leftmargin=8mm]
    \itemsep0em 
    \item \textbf{A Novel Metric and Diversity Driven Curation Strategy.} We first propose and validate Reasoning Path Divergence (RPD), a novel metric for quantifying the semantic diversity between Long-CoT solutions. Building on this metric, we propose a novel data curation pipeline. This pipeline constructs a high-quality \textit{one problem, multiple solutions} training set by selecting the most semantically distinct solutions for each problem.

   \item \textbf{Demonstrated Gains in Diversity and Performance.} Models fine-tuned on our multi-solution (1PNS) dataset achieve an average improvement of \textbf{2.80\%} in $pass@16$ performance across challenging math benchmarks, highlighted by a peak gain of \textbf{4.99\%} on the AIME24 benchmark, while simultaneously exhibiting higher output diversity as measured by our RPD metric. These gains confirm that our method alleviates the diversity bottleneck in Test-Time Scaling, boosting its overall performance.

\end{itemize}

\begin{figure}[t!]
\centering
\includegraphics[width=0.98\textwidth]{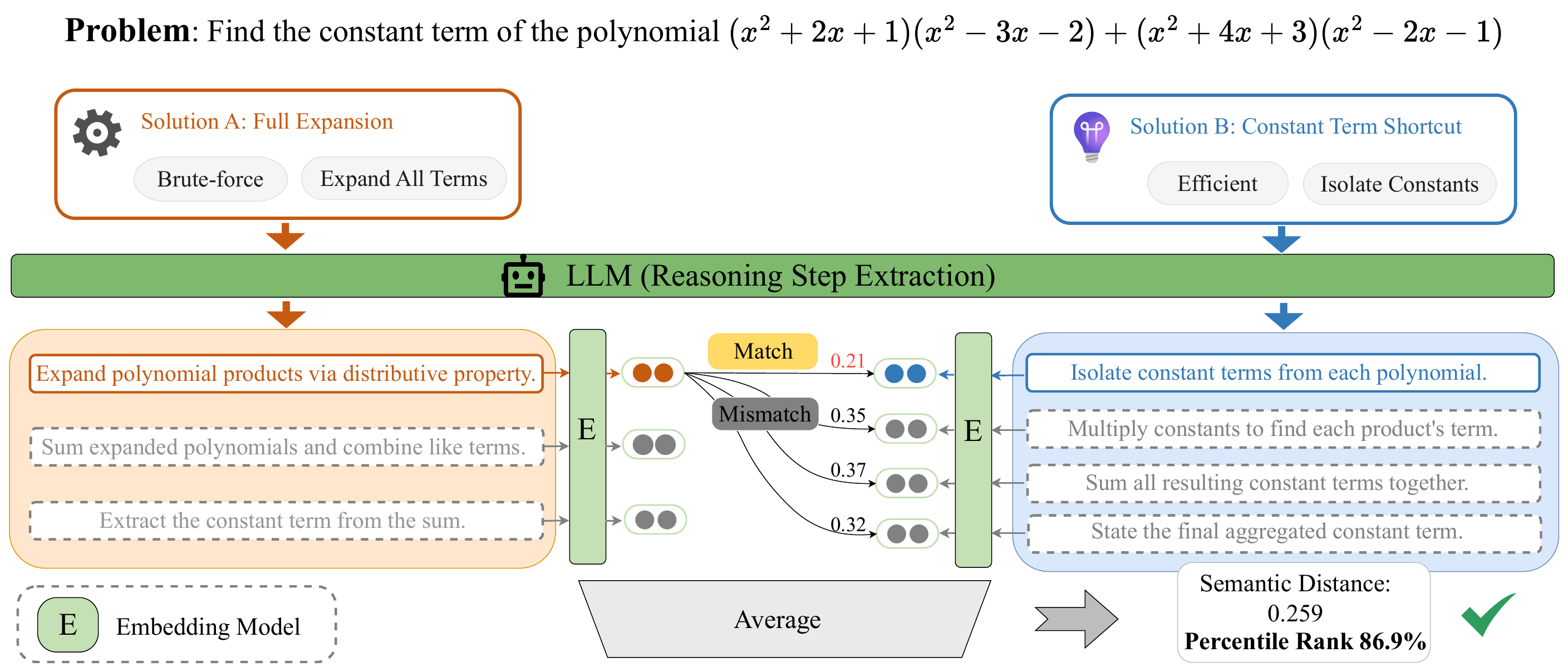}
\caption{The workflow of our Reasoning Path Divergence (RPD) metric. 
        Given two solutions (A and B), an LLM first decomposes them into step-level summaries. 
        An asymmetric matching is then performed: each step in the shorter summary (A) is matched to its semantically closest counterpart in the longer summary (B) based on embedding cosine distance. 
        The final RPD score is the average of these minimum distances. 
        Detailed examples with analysis is provided in Appendix~\ref{app:rpd_case_studies}.}
\label{fig:my_figure}
\end{figure}
\section{Related Work}
\label{sec:related work}
\paragraph{Test-Time Scaling.} A significant branch of Test-Time Scaling (TTS) focuses on improving performance by generating and aggregating multiple candidate solutions, which can be broadly divided into selection and fusion strategies. Selection-based methods choose the single best answer from a candidate pool. For example, some select the candidate with the highest verifier score, as in Best-of-N \citep{BoN1_brown,BoN2_song}, while others pick the most frequent answer via Majority Voting \citep{MV_wang}. To improve efficiency, certain studies filter candidates before final selection or voting \citep{BoN_filter_munkhbat,MV_filter_chen,MV_filter_wu}. Fusion-based methods, by contrast, merge multiple answers. This can be done by prompting an LLM to summarize the candidates \citep{fusion_jiang,fusion_li,fusion_lireasoning} or self-correct \citep{he2025selfcorrectionrefinementlearningframework}. A key challenge for these methods is low output diversity. Standard training often causes models to overfit to a single, canonical reasoning path, which limits the effectiveness of TTS.

\paragraph{LLM Generation Diversity.} A large body of work confirms that standard supervised fine-tuning is detrimental to generation diversity \citep{CE_attributing,CE_rethinking_DCO,GEM}, prompting explorations into various training-phase optimizations to mitigate this issue, especially as recent studies establish a strong positive correlation between a model's solution diversity and its reasoning potential \citep{diversity_aware_PO}. These algorithm-centric approaches are varied, ranging from modifying the training objective with techniques like confidence regularization \citep{CE_rethinking_DCO} or direct Best-of-N optimization \citep{inference_aware_finetune}, to altering the training process via sparse updates \citep{GEM}, checkpoint ensembling \citep{CE_weight_ensembling}, and lightweight, diversity-aware parameter tuning \citep{ADAPT}.  Complementing these effective, algorithm-centric strategies, our work explores a data-centric perspective aimed at directly enriching the reasoning diversity within the training data itself.

\paragraph{Data Curation.} Curating high-quality, diverse datasets is essential for effective fine-tuning \citep{albalak2024survey}. Most prior work targets \textit{inter-problem} diversity, ensuring a broad mix of distinct problems, by synthesizing new questions at scale \citep{qin2025scaling}, using automated selection frameworks \citep{liumakes}, removing semantic duplicates \citep{abbas2023semdedup}, tuning domain mixtures \citep{xie2023doremi}, or scaling environment for agentic interactions \citep{huang2025scaling}. By contrast, \textit{intra-problem} diversity, teaching multiple ways to solve the same problem, remains under-explored. Closing this gap is the aim of our work.

\section{Method}
\label{sec:method}
Enabling the \textit{one problem, multiple solutions} training paradigm requires the ability to identify and select semantically distinct reasoning paths. To address this challenge, we introduce Reasoning Path Divergence (RPD), a new fine-grained metric for Long-CoT solutions. We then describe our 1PNS Curation Pipeline, which uses RPD to systematically build a high-diversity training set from the OpenThought3 dataset \citep{openthoughtsguha2025}. This dataset contains 53,125 mathematical problems, each with 16 Long-CoT answers.

\subsection{Reasoning Path Divergence (RPD): A Step-Level Diversity Metric}
\label{sec:rpd_metric}
While prevailing approaches use embeddings over the entire solution to assess diversity, they struggle to adequately reflect solution diversity in Long-CoT reasoning, which is lengthy and structurally complex. Our RPD metric overcomes this limitation by analyzing the reasoning process at the step-summary level, computing embeddings over step summaries to capture semantic shifts in reasoning rather than superficial textual differences. For clarity, we illustrate the RPD computation framework in Figure~\ref{fig:my_figure}, which comprises two core stages:

\textbf{1. Reasoning Step Extraction.} To reduce the complexity of Long-CoT reasoning paths and enable finer-grained diversity analysis, we first decompose the Long-CoT solutions into step-wise summaries. Formally, given two Long-CoT solutions $S_A$ and $S_B$, we prompt an LLM (Qwen3-14B; \citealp{model_qwen3}) using the instructions provided in Appendix~\ref{app:summarization_prompt} to split them into a short, ordered list of step summaries: $L_A = \{a_1, ..., a_m\}$ and $L_B = \{b_1, ..., b_n\}$.

\textbf{2. Asymmetric Distance Computation.} Next, we quantify the semantic distance between the two step lists using an asymmetric matching procedure. Each step summary is encoded into a high-dimensional vector using Qwen3-Embedding-8B \citep{model_qwen3embedding}. Without loss of generality, we designate $S_A$ as the shorter solution with $m \le n$, and for each step $a_i$ in $S_A$, we identify its closest semantic match in $S_B$ by selecting the pair with the minimum cosine distance:

\vspace{-1.8em}
\begin{equation}
\begin{split}
d_i = \min_{j=1,...,n} \left(1 - \frac{\vec{e}_{a_i} \cdot \vec{e}_{b_j}}{\|\vec{e}_{a_i}\| \|\vec{e}_{b_j}\|} \right)
\end{split}
\end{equation}
%\vspace{-1.8em}

The overall RPD score, $D(S_A, S_B)$, is calculated as the average of these minimum distances:
%\vspace{-1.8em}
\begin{equation}
\begin{split}
D(S_A, S_B) = \frac{1}{m} \sum_{i=1}^{m} d_i
\end{split}
\end{equation}
\textbf{Why is RPD superior to encoding the entire sentence?} RPD’s asymmetric design is robust because it accounts for semantic diversity in summarization granularity and mitigates the impact of step order on diversity analysis. In other words, it measures the extent to which the core logic of the shorter reasoning path is captured by the longer one. As a result, if one solution is merely a more detailed restatement of another, the RPD score remains low, whereas fundamentally different reasoning strategies yield high scores. The formal definition and algorithmic details appear in Appendix~\ref{app:RPD}.

\subsection{The 1PNS Curation Pipeline}
\label{sec:curation_pipeline}

Our pipeline processes the raw OpenThought3 dataset into a high-diversity 1PNS training set in two main phases.

\textbf{Phase 1: Initial Quality Filtering.} We started with 10,000 mathematical problems from OpenThought3. Since we lacked ground-truth labels, we applied a multi-stage filtering process to ensure data quality. This included length-based filtering to set a practical \texttt{max\_new\_tokens} limit for inference, and an LLM-based screening (using Qwen3-14B) to remove unclear problems and incomplete solutions missing final answers. This first phase gave us a high-quality candidate set of 1,600 problems, each with at least 10 candidate solutions that passed the screening. More details on the protocol are in Appendix~\ref{app:quality_filtering}.

Before moving to the core selection step, we checked the natural diversity of this candidate set using a summary-based LLM Judge. As described in Appendix~\ref{app:diversity_judge_problem}, we used a Qwen3-14B model to evaluate the overall diversity of the solution summaries for each problem. The results showed a significant lack of diversity: for a majority of problems, \textbf{58\%}, all solutions followed essentially the same single reasoning strategy, with only minor differences. This finding shows that having many solutions does not automatically mean they use diverse reasoning strategies, making a dedicated problem selection phase necessary.

\textbf{Phase 2: Diversity-Driven Selection.} This phase consists of a two-stage process guided by our RPD metric:

\textbf{1. Problem Selection.} We first rank problems by their intrinsic solution diversity potential. 
Instead of averaging all pairwise distances, which can be diluted by clusters of identical solutions, we focus on detecting the existence of alternative solution approaches. 
For each problem $P$ with $k$ solutions, we compute the average divergence of each solution $S_i$ relative to all other solutions, and define the problem's overall score, $\text{Score}_{\text{div}}(P)$, as the maximum of these values:
\begin{equation}
    \text{Score}_{\text{div}}(P) = \max_{1 \le i \le k} \left( \frac{1}{k-1} \sum_{j \neq i} D(S_i, S_j) \right)
\end{equation}
We then select the top-$N$ problems from this ranked list.

\textbf{2. Solution Selection.} For each of the top-$N$ problems, we pick a set of $M$ diverse solutions. We use a greedy selection that repeatedly adds the solution with the highest average RPD relative to the solutions already chosen.

This two-stage process results in a final training set rich in strategically diverse reasoning paths. The detailed algorithm is provided in Appendix~\ref{app:curation_algorithm}.
\section{Experiments}
To validate the core hypothesis of our work—that diversity-driven data curation can enhance a model's Test-Time Scaling (TTS) performance, we designed and conducted a series of experiments. Our evaluation had two parts. First, we measured how well the Reasoning Path Divergence (RPD) metric detects strategic diversity among solutions. Second, we measured how a training set selected using RPD affected a model's downstream \texttt{pass@k} performance.

\subsection{RPD Metric Evaluation}
\label{sec:rpd_metric_eval}

\textbf{Setup.}
To evaluate RPD's effectiveness in identifying semantically diverse reasoning paths, we randomly sample 100 problems and their solutions from the high-quality candidate set from our curation pipeline (Sec.~\ref{sec:curation_pipeline}). For each problem, each method selects the pair of solutions it considers most diverse. We use a separate LLM judge to assess whether the chosen pair employs different problem-solving strategies. The success rate is reported as the main evaluation metric. The reliability of the LLM judge is validated against human annotations (see Appendix~\ref{app:exp1_llm_judge} for the full prompt and alignment study).

\textbf{Methods Compared.}
We evaluate the following methods:
\begin{itemize}
    \item \textbf{Random}: Randomly selects a pair of solutions, serving as a lower-bound baseline.
    \item \textbf{Raw Embedding (Raw Emb.)}: Selects the pair with the greatest cosine distance between the embeddings of the full solution texts.
    \item \textbf{Summary Embedding (Summary Emb.)}: Selects the pair with the greatest cosine distance between the embeddings of solution summaries.
    \item \textbf{LLM Selection}: A LLM (Qwen3-14B) selects the most diverse pair based on the summaries of all candidate solutions (see Appendix~\ref{app:exp1_llm_select} for details).
    \item \textbf{Ours (RPD)}: Our proposed asymmetric, step-level semantic distance metric.
\end{itemize}

\begin{wraptable}{r}{0.42\textwidth} 
    \vspace{-15pt} 
    \centering
    \caption{Effectiveness of various diversity metrics.}
    \label{tab:metric_results}
    \begin{tabular}{@{}lc@{}}
    \toprule
    \textbf{Method} & \textbf{Success Rate (\%)} \\ \midrule
    Random & 27 \\
    Raw Emb. & 40 \\
    LLM Selection & 44 \\
    Summary Emb. & 48 \\
    \textbf{Ours (RPD)} & \textbf{53} \\ \bottomrule
    \end{tabular}
\end{wraptable} 

\textbf{Results and Analysis.}
As shown in Table~\ref{tab:metric_results}, our RPD metric achieves a \textbf{53\%} success rate, outperforming all baselines, including those based on raw embeddings (40\%), summary embeddings (48\%), and a powerful LLM selector (44\%). These results offer two key insights. First, RPD's fine-grained, step-level analysis is crucial for overcoming the limitations of holistic embedding methods that conflate high-level strategy with superficial text. Second, RPD compares candidates in pairs systematically. This pairwise comparison is more reliable than heuristic LLM judgments for selecting the most diverse pair from many candidates. These results show that RPD works well as an automatic metric for our diversity-driven curation pipeline.

\subsection{Effectiveness of Multi-Solution Fine-Tuning}
\label{sec:main_exp}

In this experimental section, we aim to answer the following research questions:
\begin{itemize}
    \item[\textbf{Q1:}] Does fine-tuning with the one problem, multiple solutions (1PNS) paradigm improve reasoning performance, as measured by \texttt{pass@k}, compared to the standard one problem, one solution (1P1S) setting?
    
    \item[\textbf{Q2:}] Within the 1PNS paradigm, does selecting diverse solutions using our RPD metric yield higher \texttt{pass@k} than other selection methods?
\end{itemize}

\subsubsection{Experimental Setup}

\textbf{Model.} We use the Qwen3-4B-Base model \citep{model_qwen3} for our primary experiments. To ensure the robustness of our findings, corresponding results for the Qwen2.5-3B model \citep{model_qwen2.5} are provided in the Appendix~\ref{app:full_results_qwen2}, and results for Llama-3.1-8B-Instruct \citep{llama3dubey2024} are reported in Appendix~\ref{app:full_results_llama3}.

\textbf{Benchmark.} We evaluate the model's performance on three challenging mathematical reasoning benchmarks that align with our training data domain: AIME24\footnote{\url{https://huggingface.co/datasets/Maxwell-Jia/AIME_2024}}, MATH500 Level 5 \citep{dataset_math500}, and Olympiad Bench\footnote{For our evaluation, we selected an English, text-only, deterministic-answer mathematical subset of the Olympiad Bench to align with our training set.} \citep{dataset_olympiadbench}. Performance is measured using the \texttt{pass@k} metric. 

\textbf{Baselines.} 
To comprehensively evaluate our diversity-driven data curation method, we conduct a comparison against several baselines. For our main experiments, we standardize the multi-solution format to \textbf{one problem, three solutions (1P3S)}. The impact of varying the number of solutions per problem is investigated in our ablation studies (Sec~\ref{ablation}). For a fair comparison, all methods use the same total number of training instances—300.

Our proposed method, \textbf{Ours (RPD)}, constructs a training set of 100 problems and 3 solutions per problem, guided by our RPD metric's diversity scores. We compare it against the following baselines, which are grouped into two categories. The detailed construction methodology for each is provided in Appendix~\ref{app:baselines}.

\textit{1. Comparison of 1P1S vs. 1P3S paradigms.}
\begin{itemize}[leftmargin=8mm]
    \item \textbf{Random 1P1S:} The standard SFT baseline, constructed by randomly selecting 300 unique problems and pairing each with one randomly chosen solution. This baseline is used to measure the fundamental performance gain of the 1P3S approach.
\end{itemize}

\textit{2. Comparison of diversity selection metrics (all using a 1P3S structure).}
\begin{itemize}[leftmargin=8mm]
    \item \textbf{Random 1P3S:} A naive multi-solution approach. We randomly select 100 problems and use 3 randomly chosen solutions for each.
    \item \textbf{LLM Selection:} An LLM is prompted to select 100 problems and generate 3 diverse solutions for each.
    \item \textbf{Raw Embedding (Raw Emb.) :} We select the 100 problems and 3 corresponding solutions that maximize diversity based on the cosine distance between the embeddings of the full answer texts.
    \item \textbf{Summary Embedding (Summary Emb.):} We select data by maximizing the cosine distance between embeddings of AI-generated answer summaries for 100 problems and their 3 solutions.
\end{itemize}

\textit{3. Comparison against unfiltered data quantity.}
\begin{itemize}[leftmargin=8mm]
    \item \textbf{Unfiltered (1P16S):} To validate the necessity of our data curation and filtering pipeline, we implement a baseline utilizing all 16 available solutions per problem without any selection. To ensure a fair comparison controlled for the total computational budget (i.e., identical gradient update steps), we utilize a dataset of 3,600 samples (225 problems $\times$ 16 solutions) and fine-tune for 1 epoch. This matches the total training iterations of our main experiments (300 samples $\times$ 12 epochs), allowing us to determine whether diversity-driven filtering yields better results than simply maximizing data quantity.
\end{itemize}

\textbf{Implementation Details.} We fine-tune the Qwen3-4B-Base model using supervised fine-tuning with 4-bit QLoRA (rank=16, alpha=32). For our primary experiments (300 samples), the model is trained for 12 epochs. In comparisons involving larger datasets (e.g., the unfiltered baseline with 3,600 samples or scalability ablations), we adjust the training duration to 1 epoch to ensure the total number of gradient updates remains comparable. Training is conducted in BF16 precision on NVIDIA H20 GPUs. We use the AdamW optimizer with a batch size of 16 and a cosine learning rate scheduler, peaking at $5 \times 10^{-5}$. For inference, we use nucleus sampling (temperature=0.6, top\_p=0.95) with maximum generation lengths tailored to each benchmark (14K for AIME24, 10K for MATH500, 8K for Olympiad). To ensure statistical robustness, we report average scores over multiple runs (4 for AIME24/MATH500, 2 for Olympiad).

\subsubsection{Results and Analysis}

We present the experiments in two parts to answer our research questions. First, we compare the \textit{one problem multiple solutions} (1PNS) paradigm with the standard \textit{one problem one solution} (1P1S) baseline. Second, we test our diversity metric against several alternative selection strategies.

\textbf{Q1: Superiority of the 1PNS Paradigm}

\begin{figure}[t!]
    \captionsetup[subfigure]{justification=centering}
    \centering

    \begin{subfigure}[b]{0.32\textwidth}
        \centering
        \includegraphics[width=\textwidth]{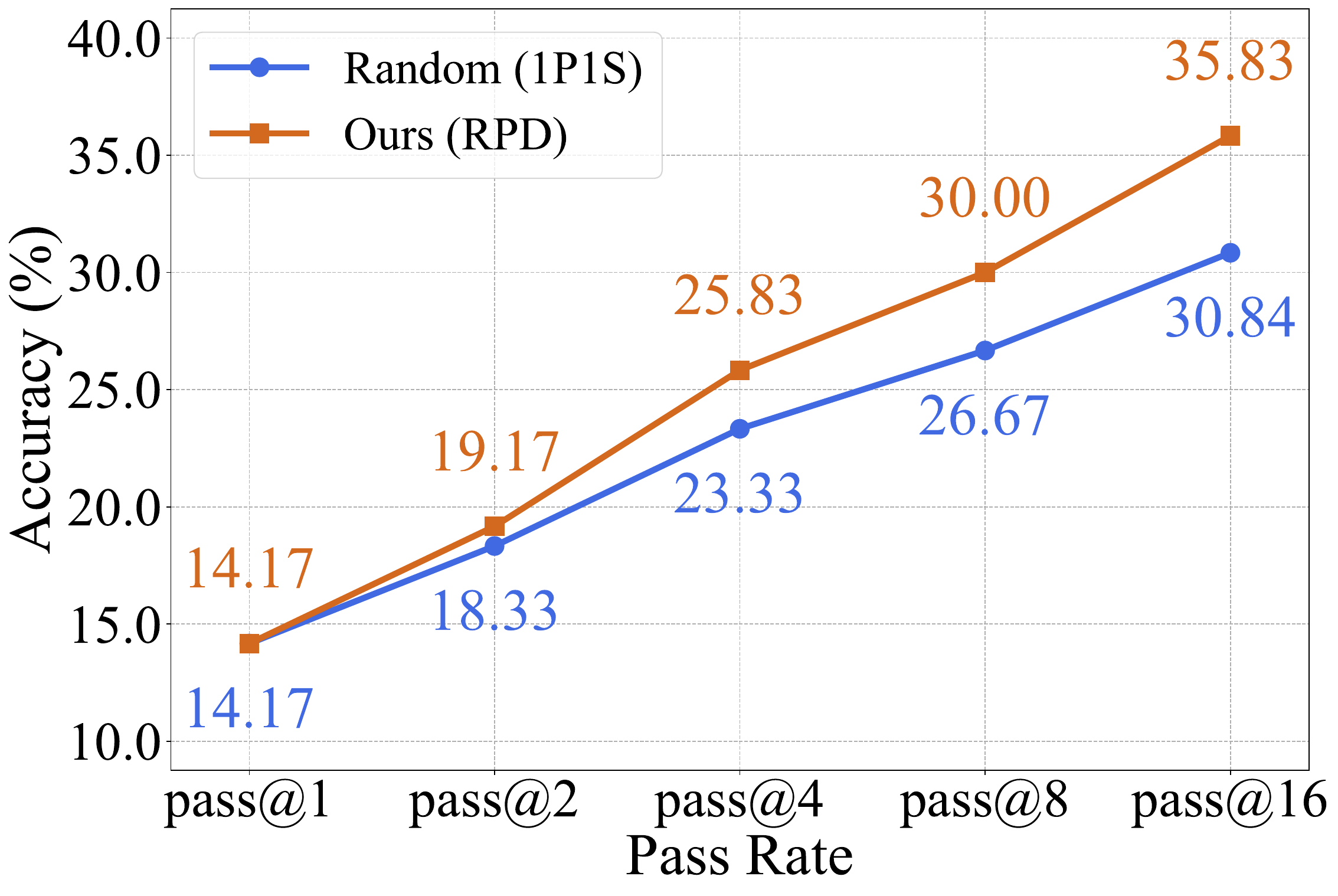}
        \caption{AIME24}
        \label{fig:aime24}
    \end{subfigure}% 
    \hfill 
    \begin{subfigure}[b]{0.32\textwidth}
        \centering
        \includegraphics[width=\textwidth]{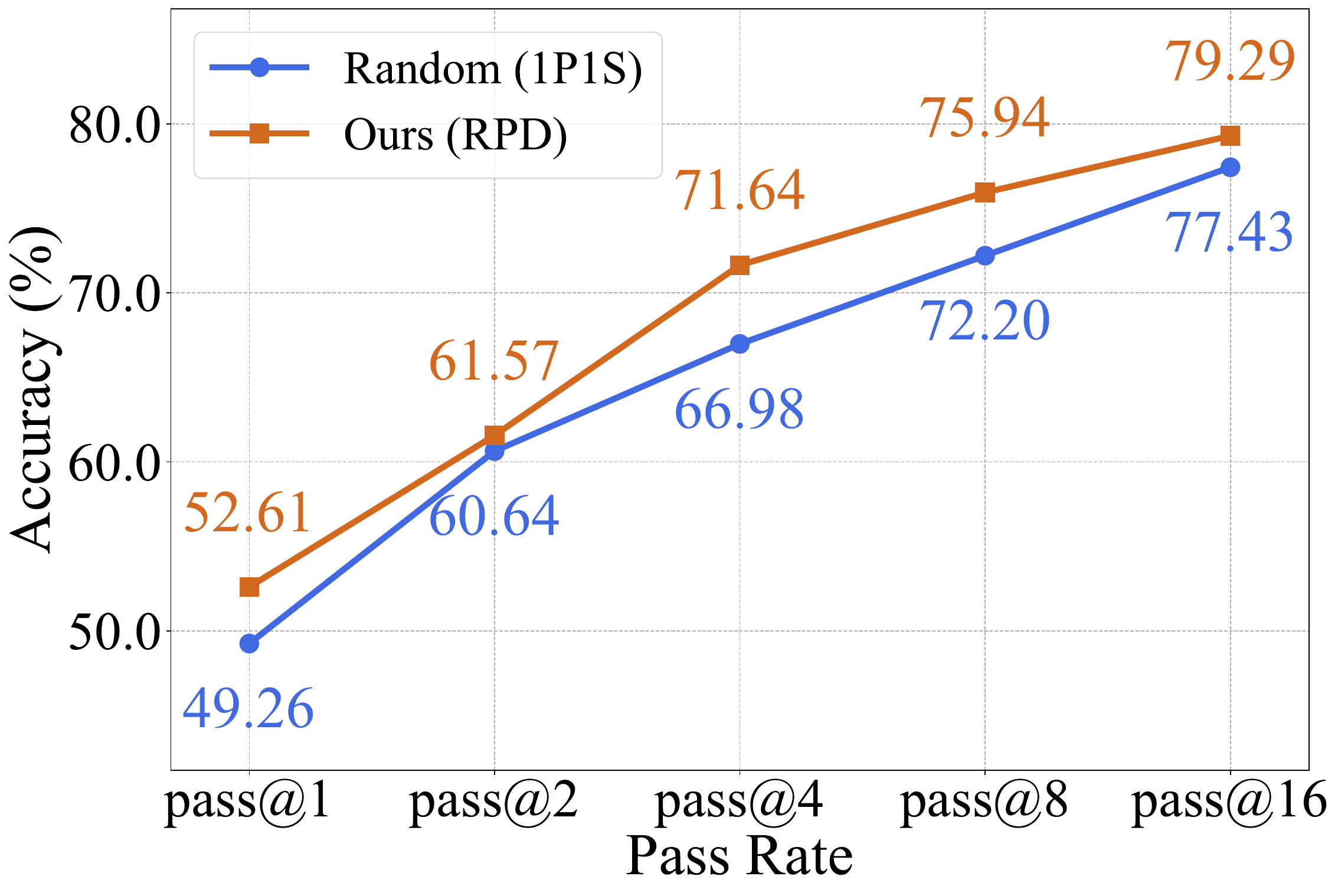}
        \caption{MATH500 Level 5}
        \label{fig:math_level5}
    \end{subfigure}% 
    \hfill 
    \begin{subfigure}[b]{0.32\textwidth}
        \centering
        \includegraphics[width=\textwidth]{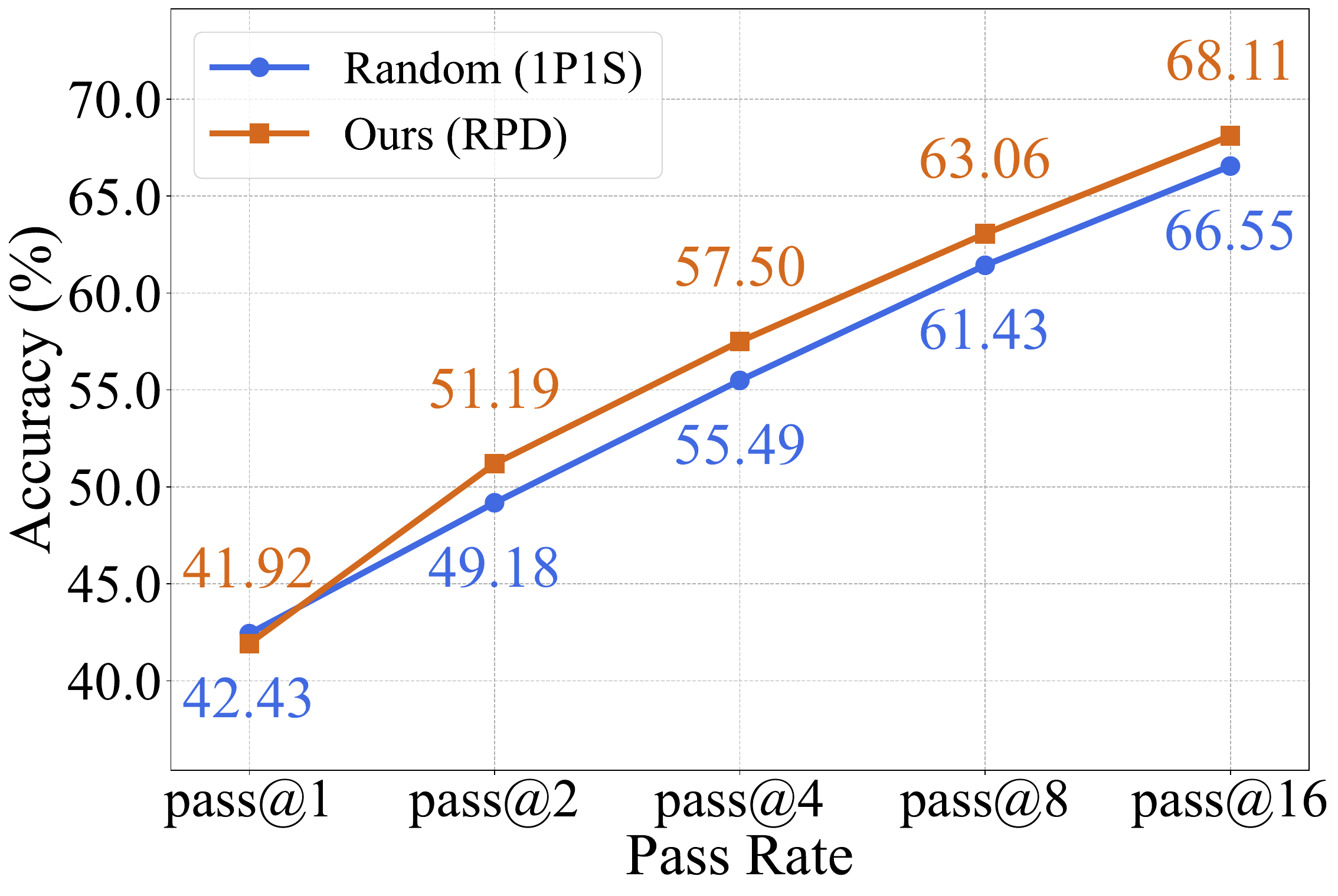}
        \caption{Olympiad Bench}
        \label{fig:olympiad_bench}
    \end{subfigure}

    \caption{Performance comparison of our 1P3S approach against the 1P1S baseline across three mathematical reasoning benchmarks. Each subplot corresponds to a different benchmark, showing the pass@k accuracy for k={1, 2, 4, 8, 16}.}
    \label{fig:main_performance_comparison}
\end{figure}

We compare our 1P3S training method to the 1P1S baseline on three benchmarks (Figure~\ref{fig:main_performance_comparison}). At \texttt{pass@1}, the two methods perform similarly. However, 1P3S outperforms the baseline for larger $k$. Across benchmarks, 1P3S achieves an average \texttt{pass@16} gain of \textbf{2.80\%}, with the largest gain of \textbf{4.99\%} on the AIME24 test. These results indicate that using multiple solutions per problem improves Test-Time Scaling on hard reasoning tasks.

\textbf{Q2: Effectiveness of the RPD Metric}

Next, we compare our diversity metric to other data selection strategies. All selection strategies use the 1P3S format (one problem, three solutions). Table~\ref{tab:diversity_comparison} shows results on the MATH500 Level 5 benchmark. Results for AIME24 and the Olympiad benchmark are in Appendix~\ref{app:full_results_qwen3_4b}.

Table~\ref{tab:diversity_comparison} shows that RPD-guided selection consistently outperforms the baseline methods on all \texttt{pass@k} metrics. Some methods, such as Summary Emb., are competitive at \texttt{pass@1}. However, our approach takes a clearer lead at larger $k$. For example, at \texttt{pass@4} our method leads by about \textbf{3.0\%} over the next-best strategy.
This gap shows that RPD better captures strategic diversity than whole-solution embedding distances or heuristic LLM selection. Finally, the fact that our method outperforms the Unfiltered (1P16S) baseline highlights the necessity of our selection pipeline, demonstrating that curated diversity is more effective than indiscriminately maximizing the quantity of solution paths.

\begin{table}[t!]
\centering
\small
\setlength{\tabcolsep}{4pt}
\caption{Comparison of different diversity selection methods on the MATH500 Level 5 benchmark. All methods except \textit{Base} use a \textbf{1P3S} structure.}
\label{tab:diversity_comparison}
\begin{tabular}{lccccc}
\toprule
\textbf{Method} & \textbf{pass@1 (\%)} & \textbf{pass@2} & \textbf{pass@4} & \textbf{pass@8} & \textbf{pass@16} \\
\midrule
\textit{Base} & 46.08 & 56.90 & 64.37 & 71.27 & 75.00 \\
Random (1P3S) & 49.07 & 59.70 & 68.66 & 73.32 & 77.24 \\
Raw Emb. & 50.19 & 59.14 & 67.54 & 71.64 & 77.80 \\
Summary Emb. & 52.24 & 59.89 & 68.66 & 73.14 & 77.43 \\
LLM Selection & 49.81 & 58.96 & 66.23 & 73.51 & 77.61 \\
Unfiltered (1P16S) & 47.20 & 57.46 & 66.60 & 73.51 & 77.80 \\
\midrule
\textbf{Ours (RPD)} & \textbf{52.61} & \textbf{61.57} & \textbf{71.64} & \textbf{75.94} & \textbf{79.29} \\
\bottomrule
\end{tabular}
\end{table}

\subsubsection{Ablation Studies}
\label{ablation}
We conduct a series of ablation studies to provide a comprehensive analysis of our method and its properties. 
% We begin by evaluating our method's impact on the diversity of generated solutions. We then analyze the sensitivity to a key hyperparameter (the number of solutions curated for each problem) before disentangling the individual contributions of our problem and answer selection components. Furthermore, we investigate whether our paradigm provides a superior foundation for subsequent reinforcement learning. Finally, we validate the scalability of our paradigm on a larger training set.

\textbf{Analysis of Solution Diversity.} 
To verify that 1PNS training increases output diversity, we analyzed 16 generated solutions for each problem in MATH Level5 test set. We partitioned problems into a \textit{moderately-solved group} (2-12 correct solutions) and a \textit{well-solved group} (13-16 correct solutions) to analyze performance on problems of varying difficulty. Diversity was measured using our RPD metric, which is the average pairwise RPD among correct solutions within each problem, and Div-Self-BLEU (100 - Self-BLEU)~\citep{self_bleu}. For both metrics, a higher score indicates greater output diversity.

Table~\ref{tab:diversity_ablation} shows that the fine-tuned model adjusts output diversity according to problem difficulty. For moderately-solved problems (2–12 correct solutions), our RPD method produces the most diverse outputs by both RPD and Div-Self-BLEU; for well-solved problems (13–16 correct solutions), diversity falls below the 1P1S baseline, suggesting convergence to a single high-confidence answer. We interpret this as an effective test-time scaling strategy: the model selectively increases exploration on challenging instances while exploiting confident solutions on simpler ones. This adaptability is key to optimizing overall \texttt{pass@k} performance.
\begin{table}[t!]
\centering
\small
\setlength{\tabcolsep}{3.5pt} 
\caption{Diversity scores for different methods on the MATH500 Level 5 test set, evaluated on the Qwen3 4B Base model. Scores are partitioned by the number of correct solutions (pass count) out of 16 attempts.}
\label{tab:diversity_ablation}
\begin{tabular}{l c c c c}
\toprule
& \multicolumn{2}{c}{\textbf{Div-Self-BLEU}} & \multicolumn{2}{c}{\textbf{Our Metric}} \\
\cmidrule(lr){2-3} \cmidrule(lr){4-5}
\textbf{Method} & \textbf{Pass Count 2-12} & \textbf{Pass Count 13-16} & \textbf{Pass Count 2-12} & \textbf{Pass Count 13-16} \\
\midrule
Random (1P1S)      & 35.27 & 15.26 & 15.17 & 13.30 \\
Random (1P3S)        & 32.52 & 14.62 & 15.57 & 14.00 \\
LLM Selection (1P3S) & 36.36 & 14.19 & 15.11 & 13.31 \\
Raw Emb. (1P3S)     & 33.94 & 14.23 & 15.39 & 13.08 \\
Summary Emb. (1P3S)  & 37.42 & 14.46 & 15.69 & 12.98 \\
Unfiltered (1P16S)  & 35.13 & 14.93 & 15.10 & 13.26 \\
\textbf{RPD (1P3S)} & 38.20 & 14.31 & 15.80 & 12.62 \\
\bottomrule
\end{tabular}
\end{table}

\textbf{Impact of the Number of Solutions per Problem.}
Keeping the training set fixed at 300 samples, we vary the number of solutions per problem and measure model performance. Table~\ref{tab:ablation_quantity} compares configurations using 2, 3, 4, or 5 solutions per problem against the single-solution baseline.

As shown in Table~\ref{tab:ablation_quantity}, the \texttt{1PNS} paradigm consistently improves performance over the single-solution baseline, and this improvement generally grows as $k$ increases. The gains peak at \textit{RPD (1P3S)} and then decline when more solutions are added, indicating a trade-off between \emph{diversity depth} and \emph{problem breadth}. The best choice depends on the dataset; for OpenThought3, the \texttt{1P3S} setting worked best in our tests.

\begin{table}[t!]
\centering
\small
\caption{Ablation study on the number of diverse solutions selected by \textbf{our RPD metric} per problem on the MATH500 Level 5 benchmark, compared against the 1P1S baseline. The total sample size is kept constant at 300.}
\label{tab:ablation_quantity}
\begin{tabular}{lccccc}
\toprule
\textbf{Configuration} & \textbf{pass@1 (\%)} & \textbf{pass@2} & \textbf{pass@4} & \textbf{pass@8} & \textbf{pass@16} \\
\midrule
Random (1P1S) & 49.26 & 60.64 & 66.98 & 72.20 & 77.43 \\
\midrule
RPD (1P2S) & 52.43 & \textbf{61.57} & 69.96 & 74.63 & 77.99 \\
\textbf{RPD (1P3S)} & 52.61 & \textbf{61.57} & \textbf{71.64} & \textbf{75.94} & \textbf{79.29} \\
RPD (1P4S)  & 52.24 & 59.70 & 70.90 & 74.63 & 79.10 \\
RPD (1P5S)  & \textbf{53.92} & 61.20 & 67.73 & 73.88 & 78.54 \\
\bottomrule
\end{tabular}
\end{table}

\textbf{Quantifying the Impact of Problem and Answer Selection Strategies.}
We isolate the effects of question selection (\texttt{RPD-Q}) and answer selection (\texttt{RPD-A}) by comparing the full method to ablations in which one component is replaced by random selection. Results are shown in Table~\ref{tab:ablation_components}. From these results, we draw three main conclusions. First, using diversity-driven selection for questions or for answers improves performance over the fully random baseline at higher \texttt{pass@k} values. Second, question selection (\texttt{RPD-Q}) has a larger effect on Test-Time Scaling than answer selection (\texttt{RPD-A}). At \texttt{pass@16}, RPD-Q yields a +1.12\% gain over random, while RPD-A alone yields +0.19\%. Finally, combining both strategies gives the best performance: at \texttt{pass@16} the full method improves by nearly one percentage point over the next-best configuration. This shows a clear synergy—Question Selection provides a strong foundation, but both components are needed to maximize reasoning performance.

\begin{table}[t!]
\centering
\small
\caption{Ablation study on the contributions of the problem (Q) and answer (A) selection components on the MATH500 Level 5 benchmark. All configurations use a 100Q, 3A structure.}
\label{tab:ablation_components}
\begin{tabular}{lccccc}
\toprule
\textbf{Method (Problem + Answer)} & \textbf{pass@1 (\%)} & \textbf{pass@2} & \textbf{pass@4} & \textbf{pass@8} & \textbf{pass@16} \\
\midrule
Random-Q + Random-A & 49.07 & 59.70 & 68.66 & 73.32 & 77.24 \\
Random-Q + RPD-A  & 50.93 & 61.38 & 68.47 & 73.69 & 77.43 \\
RPD-Q + Random-A  & 49.81 & 59.52 & 67.91 & 74.82 & 78.36 \\
\midrule
\textbf{RPD-Q + RPD-A (Ours)} & \textbf{52.61} & \textbf{61.57} & \textbf{71.64} & \textbf{75.94} & \textbf{79.29} \\
\bottomrule
\end{tabular}
\end{table}

\textbf{Effect of Subsequent Reinforcement Learning}
\label{ablation:rl}
To evaluate whether the diverse reasoning trajectories encouraged by our 1PNS paradigm yield a better foundation for subsequent reinforcement learning (RL) fine-tuning, we applied an additional RL phase on both the RPD-curated (1P3S) model and the Random (1P1S) baseline using the historical AIME problems (1983--2023)\footnote{\url{https://huggingface.co/datasets/gneubig/aime-1983-2024}} (see Appendix~\ref{app:rl_details} for experimental details).

Table~\ref{tab:ablation_rl} summarizes the results. RL fine-tuning improves performance for both models across the board. However, the model initialized with RPD-selected data consistently outperforms the 1P1S baseline on every benchmark and for every evaluation metric (\texttt{pass@1/2/4/8/16}). Notably, the advantage in \texttt{pass@1} (e.g., surpassing the baseline by 4.47 points on MATH500) suggests that our initialization establishes a more robust capability prior to RL.

These findings support the hypothesis that SFT with diverse, RPD-selected reasoning paths provides a stronger starting point for RL, enabling larger downstream gains.

\begin{table}[t!]
\centering
\small
\setlength{\tabcolsep}{3pt} 
\caption{Performance comparison after an additional phase of Reinforcement Learning (RL) fine-tuning. The models were first fine-tuned with SFT (Random 1P1S vs. RPD 1P3S) and then further tuned with RL on the Simple-Zoo dataset.}
\label{tab:ablation_rl}
\begin{tabular}{llccccc}
\toprule
\textbf{Benchmark} & \textbf{Method} & \textbf{pass@1 (\%)} & \textbf{pass@2} & \textbf{pass@4} & \textbf{pass@8} & \textbf{pass@16} \\
\midrule
\multirow{2}{*}{AIME24} 
& Random (1P1S) + RL & 15.83 & 20.00 & 25.00 & 28.34 & 32.50 \\
& RPD (1P3S) + RL & \textbf{17.50} & \textbf{21.67} & \textbf{27.50} & \textbf{32.50} & \textbf{36.67} \\
\midrule
\multirow{2}{*}{MATH500 Level 5} 
& Random (1P1S) + RL & 56.72 & 62.69 & 68.66 & 74.63 & 79.10 \\
& RPD (1P3S) + RL & \textbf{61.19} & \textbf{69.40} & \textbf{73.89} & \textbf{77.61} & \textbf{82.09} \\
\midrule
\multirow{2}{*}{Olympiad Bench} 
& Random (1P1S) + RL & 45.99 & 53.41 & 59.94 & 66.62 & 69.44 \\
& RPD (1P3S) + RL & \textbf{47.77} & \textbf{56.38} & \textbf{62.76} & \textbf{67.06} & \textbf{71.07} \\
\bottomrule
\end{tabular}
\end{table}

 \textbf{Scalability to Larger Datasets.}
Our strategy retains its superiority over the 1P1S baseline when scaled to 3,000 samples (Appendix~\ref{app:scale_up_ablation}).

\textbf{Computational Efficiency.}
We further profiled the runtime cost in Appendix~\ref{app:compute_overhead}. We emphasize that this is a one-time dataset construction process. Thus, the curation overhead (approx. 1.68s per solution) is well-justified given the long-term reusability of the data.

\textbf{Robustness of Metric Calculation.}
We also investigated pipeline stability using a smaller summarizer. As shown in Appendix~\ref{app:summarizer_ablation}, our method maintains its effectiveness even with a 7B model, demonstrating that RPD is robust and not dependent on specific large-scale models.

\textbf{Generalization to Code Generation.}
Extending evaluations to the code domain (Appendix~\ref{app:code_generalization}) shows consistent improvements, confirming the 1PNS paradigm's applicability beyond mathematics.
\section{Conclusion}
\label{sec:conclusion}
To enable the \textit{one problem, multiple solutions} (1PNS) paradigm, we introduce a novel metric for quantifying reasoning diversity, Reasoning Path Divergence (RPD), and use it to curate a dataset of maximally diverse solutions. Our experiments validate the superiority of the 1PNS paradigm over the standard 1P1S baseline: training on RPD-curated data reduces output homogenization and yields significant \texttt{pass@k} gains. These results show that our approach provides a direct way to improve test-time scaling.

\clearpage

\section*{Acknowledgements}
This research was carried out within the HKUST Summer Research Internship Program, a collaboration between the Hong Kong University of Science and Technology (HKUST) and the University of Science and Technology of China (USTC). The authors would like to thank the Ren.AI Lab at HKUST for access to computational resources and research facilities. We also thank Mr. Zhiyuan Fan, an MPhil student at HKUST, for helpful comments during the early stage of this work.

\bibliographystyle{abbrvnat}
\bibliography{custom}
\clearpage
\appendix

\section{RPD  Curation Method Implementation}
\subsection{Step-wise Solution Summarization via LLM}
\label{app:summarization_prompt}

Our proposed diversity metric relies on a fine-grained, step-by-step summary of the reasoning path for each solution. To create these summaries, we use an LLM (Qwen3-14B) to break down each solution into its core logical steps. A key challenge is to ensure these summaries accurately reflect the original methodology while maintaining a consistent level of granularity. Overly concrete summaries might capture superficial numerical differences, while overly abstract summaries might fail to distinguish between genuinely different strategies.

To solve this, we design a detailed prompt that controls the LLM's output format and level of abstraction. This prompt instructs the model to produce a structured JSON object containing 3 to 5 method-focused steps. This strict format helps maintain uniformity across all summarized solutions. The complete prompt is provided below.

\begin{tcolorbox}[
    breakable,
    colback=gray!10!white,
    colframe=gray!75!black,
    title=Prompt for Step-wise Solution Summarization,
    fonttitle=\bfseries,
    arc=2mm,
    boxrule=1pt,
]

% --- Introductory text ---
You are a specialized AI expert in analyzing mathematical solutions. Your task is to first provide a step-by-step analysis of a solution, and then, based on your analysis, generate a final JSON output that is concise, direct, and method-focused.

\subsubsection*{REQUIRED OUTPUT STRUCTURE}
Your response \textbf{MUST} have two distinct parts in the following order:

\paragraph{Part 1: Analysis \& Thinking Process}
\begin{itemize}
    \item Start this section with the heading \texttt{\#\#\# Analysis}.
    \item Briefly explain your reasoning as you deconstruct the provided solution. This is your ``scratchpad".
\end{itemize}

\paragraph{Part 2: Final JSON Output}
\begin{itemize}
    \item After your analysis, provide the final JSON output enclosed in \texttt{//boxed\{\{\}\}}.
    \item This part must contain \textit{only} the \texttt{//boxed\{\{...\}\}} block and nothing else.
\end{itemize}

\subsubsection*{CONTENT RULES FOR THE FINAL JSON}
\begin{enumerate}[label=\arabic*.]
    \item \textbf{Step Count}: The JSON must contain \textbf{strictly 3 to 5 logical steps}.
    \item \textbf{Output Style}:
    \begin{itemize}
        \item \textbf{Use direct, active verb phrases.} Start each description with a verb (e.g., ``Calculate", ``Identify", ``Apply").
        \item \textbf{DO NOT use narrative phrasing} like ``The author identifies..." or ``The solution then calculates...".
    \end{itemize}
    \item \textbf{Abstraction Level}:
    \begin{itemize}
        \item Be abstract about numbers and variables, but \textbf{be specific about the methodology}.
        \item \textbf{BAD (Too Vague):} ``Use a formula to get the result."
        \item \textbf{BAD (Too Concrete):} ``Calculate 1/3 + 1/6 = 1/2."
        \item \textbf{GOOD (Balanced):} ``Combine the individual rates to find the total work rate."
    \end{itemize}
\end{enumerate}

\subsubsection*{JSON STRUCTURE SPECIFICATION}
\begin{itemize}
    \item The root object must have one key: \texttt{"logical\_steps"}.
    \item The value of \texttt{"logical\_steps"} must be a list (\texttt{[]}) of step objects.
    \item Each step object (\texttt{\{\{\}\}}) must contain two keys:
    \begin{itemize}
        \item \texttt{"step\_title"}: A short title for the step (e.g., ``Step 1: Combine Rates"). Use \texttt{null} if not applicable.
        \item \texttt{"step\_description"}: A concise summary of the action, following all rules above.
    \end{itemize}
\end{itemize}

\subsubsection*{EXAMPLE OF THE COMPLETE TWO-PART OUTPUT}
\textbf{Input Solution}: ``Pipe A fills a tank in 3 hours, so its rate is 1/3 tank/hr. Pipe B fills it in 6 hours, so its rate is 1/6 tank/hr. Together, their rate is 1/3 + 1/6 = 1/2 tank/hr. Therefore, the time to fill the tank together is the reciprocal of the rate, which is 1 / (1/2) = 2 hours."

\vspace{1ex}
\textbf{Your Required Output}:
\begin{lstlisting}[breaklines=true, basicstyle=\ttfamily]
### Analysis
The solution addresses a classic work-rate problem.
1.  First, it calculates the individual rate for each pipe.
2.  Second, it sums these rates to get a combined rate.
3.  Finally, it converts the combined rate back into total time.
The logic is broken down into three clear, abstract steps.

//boxed{{
  "logical_steps": [
    {{
      "step_title": "Step 1: Determine Individual Rates",
      "step_description": "Determine the individual work rate of each component based on the time taken."
    }},
    {{
      "step_title": "Step 2: Combine Rates",
      "step_description": "Combine the individual rates to find the total system work rate."
    }},
    {{
      "step_title": "Step 3: Calculate Total Time",
      "step_description": "Calculate the total time by taking the reciprocal of the combined work rate."
    }}
  ]
}}
\end{lstlisting}
\hrulefill
\vspace{1em}

\subsubsection*{YOUR TASK}
\textbf{Math Problem}:\\
\texttt{\{question\_text\}}

\vspace{1em}
\textbf{Chain-of-Thought Solution to Analyze}:\\
\texttt{\{answer\_cot\}}

\end{tcolorbox}

\subsection{Reasoning Path Divergence (RPD) Calculation}
\label{app:RPD}
After summarizing each solution into a series of core logical steps, the next phase is to compute the pairwise diversity using our \textbf{Reasoning Path Divergence (RPD)} metric. RPD is designed to quantify the semantic distance between the step-lists of two solutions, $S_A$ and $S_B$.

The calculation begins by embedding each logical step using the \textbf{Qwen3-Embedding-8B} model. Subsequently, it computes an asymmetric score by finding the average minimum cosine distance from the steps of the shorter solution to all steps in the longer one. This asymmetric design is crucial: it ensures that a solution containing a genuinely novel step is considered distant, even if its other steps are subsumed by a more comprehensive solution. The formal algorithm is detailed below.

\begin{algorithm}[H]
\caption{Reasoning Path Divergence (RPD) Calculation}
\label{alg:diversity_metric_robust}
\begin{algorithmic}[1]
    \Require Two Long-CoT solutions, $S_A$ and $S_B$.
    \Ensure A scalar diversity score $D \in [0, 1]$.
    \State $L_A \gets \text{ExtractSteps}(S_A)$; $L_B \gets \text{ExtractSteps}(S_B)$
    \If{$L_A \text{ is empty or } L_B \text{ is empty}$} 
        \State \textbf{return} $1.0$
    \EndIf
    \Statex
    
    \State $E_A \gets \{ \text{Embed}(a_i) \mid a_i \in L_A \}$; $E_B \gets \{ \text{Embed}(b_j) \mid b_j \in L_B \}$
    \Statex
    
    \State $(E_{\text{shorter}}, E_{\text{longer}}) \gets 
        \begin{cases} 
            (E_A, E_B) & \text{if } |E_A| \le |E_B| \\
            (E_B, E_A) & \text{otherwise}
        \end{cases}$
    \Statex
    
    \State $\text{min\_distances} \gets \emptyset$
    \ForAll{$\vec{e}_s \in E_{\text{shorter}}$}
        \State $d_{\min} \gets \min_{\vec{e}_l \in E_{\text{longer}}} \left(1 - \frac{\vec{e}_s \cdot \vec{e}_l}{\|\vec{e}_s\| \|\vec{e}_l\|}\right)$
        \State $\text{min\_distances} \gets \text{min\_distances} \cup \{d_{\min}\}$
    \EndFor
    \Statex
    
    \State $D_{\text{final}} \gets \text{Mean}(\text{min\_distances})$
    \State \textbf{return} $D_{\text{final}}$
\end{algorithmic}
\end{algorithm}

\clearpage
\subsection{Diversity-Driven Data Curation}
\label{app:curation_algorithm}
Our data curation process is a two-stage procedure designed to build a training set rich in strategic diversity. First, we perform \textbf{Problem Selection} to identify problems where distinct reasoning strategies are present by scoring each problem based on its intrinsic diversity potential. Second, for each of these top-ranked problems, we execute a greedy \textbf{Solution Selection} algorithm to curate a small but maximally diverse subset of $M$ solutions. This two-stage approach ensures both inter-problem and intra-problem diversity. The algorithms for both stages are detailed below.

\begin{algorithm}[H]
\caption{Stage 1: Problem selection by intrinsic diversity}
\label{alg:problem_selection_compact}
\begin{algorithmic}[1]
    \Require Candidate problem set $\mathcal{P}$, target count $N$, pairwise distance function $D(\cdot,\cdot)$
    \Ensure Top-$N$ problems $\mathcal{P}_{\text{top}}$ ranked by intrinsic diversity
    \State Initialize empty list of pairs $\mathcal{L}\gets[]$
    \ForAll{problem $P\in\mathcal{P}$}
        \State Let $\mathcal{S}_P=\{S_1,\dots,S_{k_P}\}$ be its candidate solutions
        \If{$k_P < 2$}
            \State append $(P, -\infty)$ to $\mathcal{L}$ 
            \State \textbf{continue}
        \EndIf
        \State $\text{score} \gets \max_{1 \le i \le k_P} \left( \dfrac{1}{k_P-1} \sum_{j \neq i} D(S_i, S_j) \right)$
        \State append $(P,\ \text{score})$ to $\mathcal{L}$
    \EndFor
    \State Sort $\mathcal{L}$ by score (second element) in descending order
    \State $\mathcal{P}_{\text{top}} \gets$ first $\min(N,|\mathcal{P}|)$ problems from sorted $\mathcal{L}$
    \State \Return $\mathcal{P}_{\text{top}}$
\end{algorithmic}
\end{algorithm}

\begin{algorithm}[h]
\caption{Stage 2: Greedy Selection}
\label{alg:greedy_maxmin_compact}
\begin{algorithmic}[1]
    \Require Candidate solutions $\mathcal{S}_{\text{cand}}=\{S_1,\dots,S_k\}$, pairwise distance matrix $\mathbf{D}\in\mathbb{R}^{k\times k}$, target size $M$
    \Ensure Selected index set $\mathcal{I}_{\text{select}}$ with $|\mathcal{I}_{\text{select}}|=\min(M,k)$
    \If{$M \le 0$ \textbf{or} $k = 0$} \Return $\emptyset$ \EndIf
    \If{$M \ge k$} \Return $\{1,\dots,k\}$ \EndIf

    \State $i_{\text{first}} \gets \arg\max_{i}\sum_{j\ne i}\mathbf{D}_{ij}$ 
    \State $\mathcal{I}_{\text{select}} \gets \{i_{\text{first}}\}$;\quad $\mathcal{I}_{\text{remain}}\gets\{1,\dots,k\}\setminus\{i_{\text{first}}\}$
    \State \textbf{for each} $r\in\mathcal{I}_{\text{remain}}$ set $m[r]\gets \mathbf{D}_{r,i_{\text{first}}}$ 

    \While{$|\mathcal{I}_{\text{select}}| < M$ and $\mathcal{I}_{\text{remain}}\neq\emptyset$}
        \State $r^\star \gets \arg\max_{r\in\mathcal{I}_{\text{remain}}} m[r]$ 
        \State $\mathcal{I}_{\text{select}}.\text{append}(r^\star)$; \quad $\mathcal{I}_{\text{remain}}.\text{remove}(r^\star)$
        \State \textbf{for each} $r\in\mathcal{I}_{\text{remain}}$: $m[r] \gets \min\big(m[r],\,\mathbf{D}_{r,r^\star}\big)$ 
    \EndWhile

    \State \Return $\mathcal{I}_{\text{select}}$
\end{algorithmic}
\end{algorithm}

\clearpage
\section{Dataset Preprocessing and Analysis}
\subsection{Detailed Dataset Filtering Protocol}
\label{app:quality_filtering}

The OpenThought3 dataset is a valuable open-source resource, containing approximately 53,000 mathematical problems, each with 16 corresponding completions. However, the raw dataset presents several challenges for direct use in supervised fine-tuning. Key issues include the absence of ground truth labels, the possibility of encountering ambiguous or ill-posed problems, and the fact that some solutions may be unfinished or lack a definitive final answer. Furthermore, the length of the provided solutions varies dramatically.

To curate a high-quality training corpus and ensure computational efficiency during model inference, we implement a rigorous two-stage filtering protocol on a subset of 10,000 problems from OpenThought3. This protocol addresses both solution length and quality.

\paragraph{Stage 1: Length-Based Filtering.}
Our first step is to control for solution length. This measure is primarily motivated by the practical need to set a reasonable \texttt{max\_new\_tokens} parameter during inference. Accordingly, we filter out any problem whose average token count across all its solutions exceeds 14,000 tokens.

\paragraph{Stage 2: Quality and Completeness Filtering.}
Next, we address the issue of solution quality and completeness. We employ an LLM (Qwen3-14B) as a judge to verify whether each solution is valid. For every solution in the length-filtered set, we provide its final 500 tokens as input to the LLM. The model is instructed to determine if the solution concludes properly by presenting a clear and final answer. Solutions that the LLM judge flags as incomplete or inconclusive are discarded, and any problem subsequently left with fewer than 10 valid solutions is also removed.

This comprehensive filtering pipeline refines the initial pool of 10,000 problems into a high-quality, curated set of \textbf{approximately 1,600 problems}. Each problem in this final set has an average solution length of less than 14,000 tokens and is accompanied by at least 10 complete, validated solutions. This curated 1,600-problem dataset serves as the foundation for all subsequent experiments conducted in this work.
\subsection{Dataset Diversity Analysis}
\label{app:diversity_judge_problem}

To better inform our data curation, we first analyze the existing strategic diversity within our high-quality candidate set. We use a summary-based LLM Judge to classify whether the solutions for each problem are strategically uniform or diverse.

For each problem, we concatenate the step-wise summaries of all its candidate solutions (detailed in Appendix~\ref{app:summarization_prompt}) into a single string. This, along with the original problem statement, is provided to an LLM Judge (Qwen3-14B). The judge's task is to perform a binary classification on the entire set of solutions, identifying if at least two different solution strategies are present.

We specifically write the prompt to instruct the model to ignore superficial differences in wording or calculation, and instead focus on fundamental strategic choices, such as using direct casework versus complementary counting. We do this so that the classification reflects genuine methodological diversity, not just surface-level variations. The insights from this analysis, as reported in the main text, confirm the need for our subsequent diversity-driven problem selection phase. The complete prompt for this task is detailed below.

\begin{tcolorbox}[ 
    breakable, 
    colback=gray!10!white,   % Light gray background 
    colframe=gray!75!black,  % Dark gray frame 
    title=Prompt for Problem Classification, 
    fonttitle=\bfseries, 
    arc=2mm, 
    boxrule=1pt 
] 

You are a master mathematician and an expert in pedagogical analysis. Your task is to classify a problem based on the methodological diversity of its proposed solutions. 

Your goal is to perform a binary classification: 
\begin{itemize} 
    \item \textbf{Class 2 (Diverse):} If there are at least two distinct core methodologies present across all the provided solution summaries. 
    \item \textbf{Class 1 (Not Diverse):} If all solutions use the same core methodology, or if the differences are only superficial (e.g., a different order of calculation, or using standard procedural equivalents like substitution vs. elimination). 
\end{itemize} 

\hrulefill 
\subsubsection*{1. Your Analysis Framework \& Core Criteria} 
\hrulefill 

Your primary task is to act as a discerning analyst. You must distinguish between minor procedural choices and significant differences in core steps. Assume that most solutions might share a high-level strategy; your goal is to find answers that execute core steps in a meaningfully different way. 

\paragraph{Defining Methodological Difference (Your Core Criteria):} 

\paragraph{\textbf{What IS NOT a Significant Difference (Methodologically Similar):}} 
\begin{itemize} 
    \item \textbf{Order of Calculation:} Calculating value A then B, versus B then A, before combining them in the same way. 
    \item \textbf{Algebraic Equivalence:} Using the form $(a+b)^2$ versus $a^2 + 2ab + b^2$. 
    \item \textbf{Variable Naming or Notation:} Using $n$ vs $x$. 
    \item \textbf{Choice of Standard Procedural Equivalents:} One summary describes solving a system of equations using \textbf{substitution}, while the other uses \textbf{elimination}. These are considered standard, interchangeable procedures within the same overall algebraic approach. 
    \item \textbf{Rigorous Proof vs. Heuristic Assumption:} If the overall strategy is the same, simply proving a result versus assuming it does not constitute a diverse approach. Both are still following the same high-level logical path. 
\end{itemize} 

\paragraph{\textbf{What IS a Significant Difference (Methodologically Diverse):}} 
\begin{itemize} 
    \item This difference represents a \textbf{completely distinct, independent, high-level strategic choice} that fundamentally alters the entire problem-solving path from beginning to end. 
    \item \textbf{Example 1 (Different Overall Framework):} One solution to a geometry problem uses \textbf{coordinate geometry}, another uses \textbf{synthetic geometry}, and a third uses \textbf{vector analysis}. 
    \item \textbf{Example 2 (Completely Different Logical Path):} To solve a counting problem, one answer uses \textbf{direct casework}, another uses \textbf{complementary counting}, and a third uses a \textbf{recurrence relation}. 
    \item \textbf{Example 3 (Change in Analytical Tool):} A solution to an optimization problem uses \textbf{calculus}, a second uses \textbf{inequalities} (like AM-GM), and a third uses \textbf{linear programming}. 
\end{itemize} 

\hrulefill 
\subsubsection*{2. Content to Analyze} 
\hrulefill 

\textbf{Problem:}\\ 
\texttt{\{question\}} 

\vspace{1em} 
\textbf{Proposed Solutions (Summarized by Logical Steps):}\\ 
\texttt{\{summaries\_text\}} 

\hrulefill 
\subsubsection*{3. Output Requirement} 
\hrulefill 

Based on the final criteria review, classify the diversity of the solutions. 

\textbf{Output Requirement:}\\ 
Immediately after your classification, provide your final answer in a strict JSON format within a special block. The JSON should be a single integer, either \texttt{1} or \texttt{2}. Do not provide any other text. 

\vspace{0.5em} 
Example of Final Output Structure for a \textbf{Diverse} problem: 
\begin{verbatim} 
//boxed{{2}} 
\end{verbatim} 

Example of Final Output Structure for a \textbf{Not Diverse} problem: 
\begin{verbatim} 
//boxed{{1}} 
\end{verbatim} 

Begin Analysis and Provide Output: 

\end{tcolorbox}

This classification process is applied to the 1,600 high-quality problems in our candidate pool, yielding the diversity distribution statistics reported in Section~\ref{sec:curation_pipeline}.

\clearpage

\section{Case Studies and Analysis of the RPD Metric}
\label{app:rpd_case_studies}

To provide a deeper insight into the effectiveness of our RPD metric, this section presents both a statistical overview and concrete, illustrative examples comparing it against a standard baseline.

\subsection{Statistical Distribution of Diversity Scores}

We first analyze the overall behavior of RPD compared to a common baseline. The baseline method calculates the cosine distance between the embeddings of the full, raw solution texts. We sampled 100 problems from our candidate pool and computed all pairwise diversity scores for their solutions using both methods, resulting in a total of 8,986 data points (i.e., solution pairs) for each distribution.

\begin{figure}[H]
    \centering
    \includegraphics[width=0.48\textwidth]{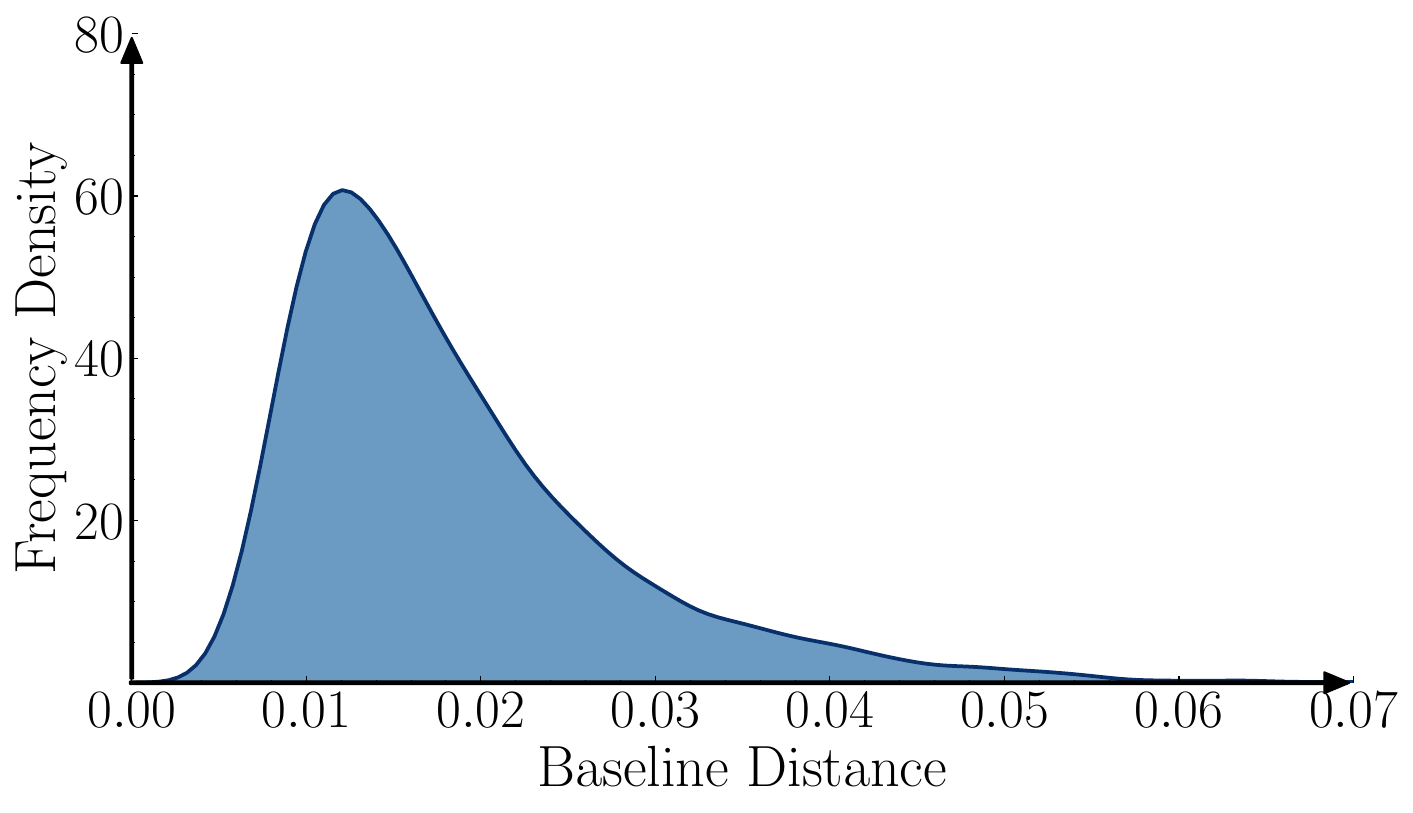}
    \includegraphics[width=0.48\textwidth]{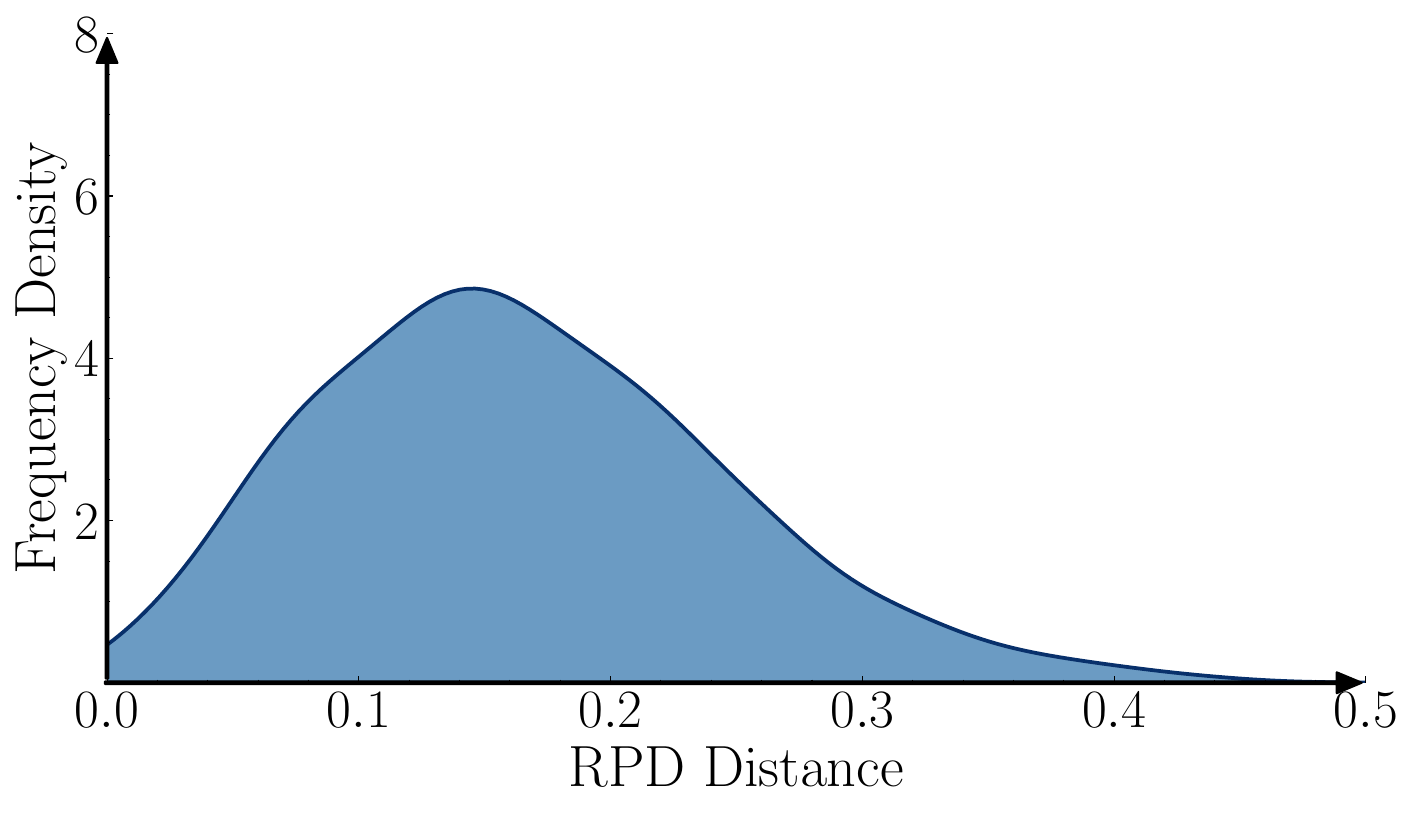}
    \caption{Distribution of pairwise diversity scores on 100 problems for the baseline (left) and our RPD metric (right). RPD provides a significantly better-separated distribution.}
    \label{fig:score_distribution}
\end{figure}

Figure~\ref{fig:score_distribution} illustrates the resulting score distributions. The baseline scores are heavily concentrated in a very narrow range near zero (0.00--0.04). This indicates that full-text embeddings are largely insensitive to the underlying reasoning structure, assigning nearly identical low-diversity scores to most pairs and failing to distinguish between subtle and significant strategic differences. In contrast, our RPD metric produces a much wider and more uniform distribution. This indicates that RPD possesses significantly higher resolution and sensitivity, allowing it to capture a continuous spectrum of strategic differences, from the subtle to the substantial.

\clearpage
\subsection{Illustrative Examples}
The following case studies provide concrete examples of this phenomenon.

\begin{figure}[h!]
  \centering
  \begin{subfigure}[t]{0.49\textwidth}
    \centering
    \includegraphics[width=\linewidth]{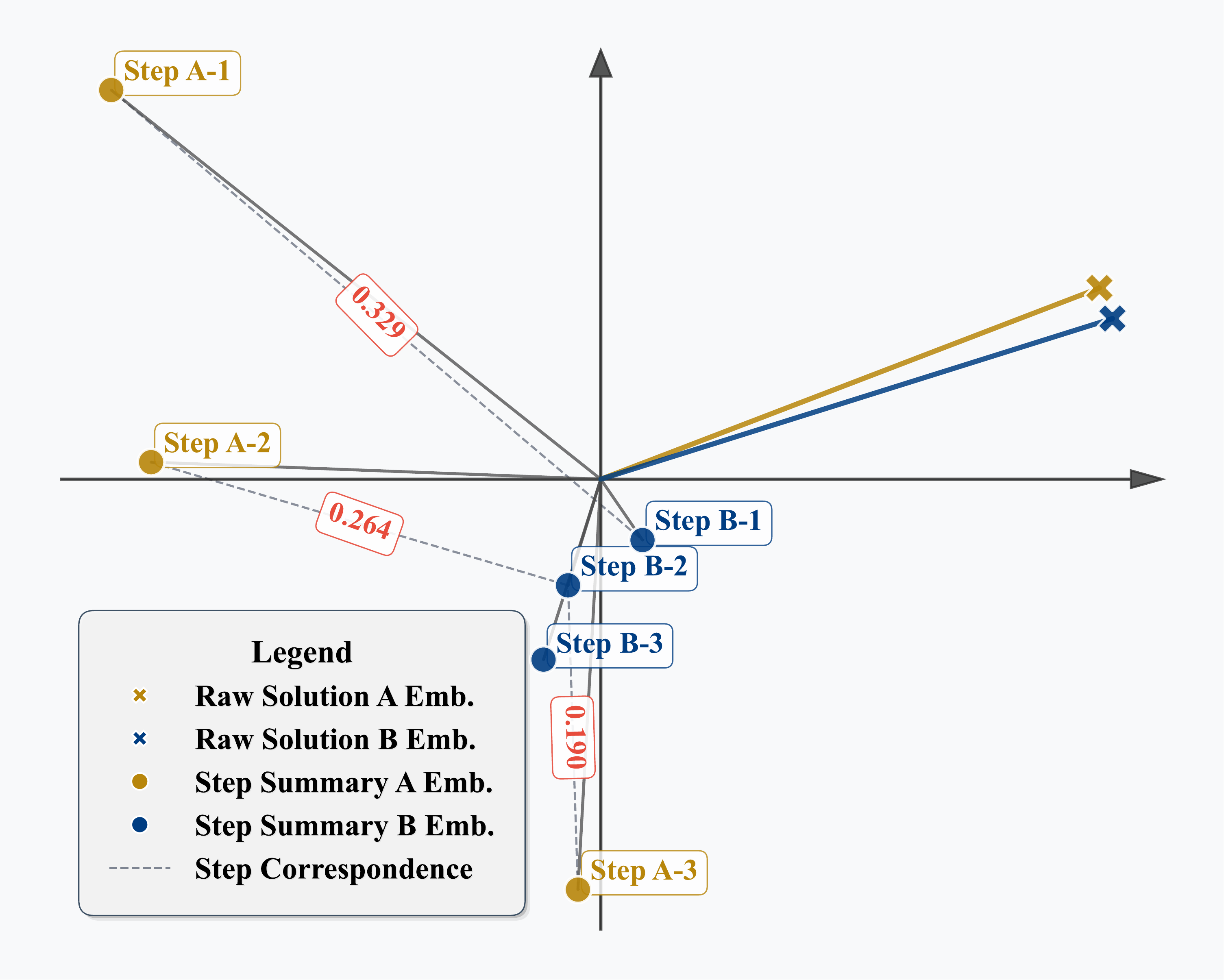}
    \caption[Case Study 1]{Case Study 1\\\small Raw Emb. Distance: 0.015 (Percentile: 44.46\%)\\\small RPD Distance: 0.259 (Percentile: 86.92\%)}
    \label{fig:pca_example_1}
  \end{subfigure}\hfill
  \begin{subfigure}[t]{0.49\textwidth}
    \centering
    \includegraphics[width=\linewidth]{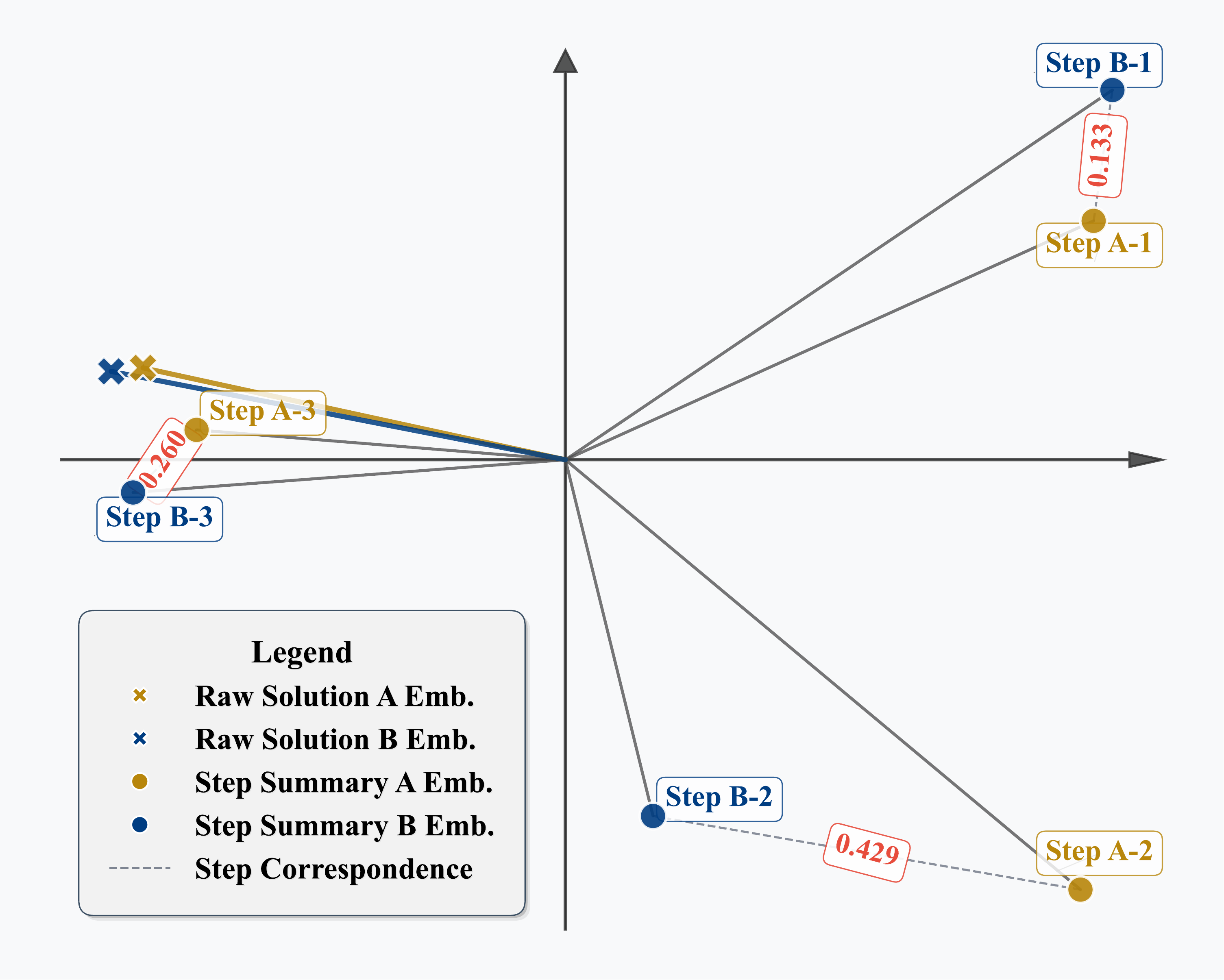}
    \caption[Case Study 2]{Case Study 2\\\small Raw Emb. Distance: 0.016 (Percentile: 52.14\%)\\\small RPD Distance: 0.274 (Percentile: 90.44\%)}
    \label{fig:pca_example_2}
  \end{subfigure}
  \caption{PCA visualization of raw solution and step summary embeddings. The step embeddings for the two solutions occupy distinct regions of the space, reflecting a strategic diversity that our RPD metric correctly identifies. In contrast, the raw solution embeddings are nearly collinear, causing the baseline method to fail to distinguish them.}
  \label{fig:pca_plots}
\end{figure}

\textbf{Case Study 1: Summaries for Figure~\ref{fig:pca_example_1}}
\begin{itemize}
    \item \textbf{Question:} Find the constant term in the polynomial $(x^2 + 2x + 1)(x^2 - 3x - 2) + (x^2 - 2x - 1)(x^2 + 4x + 3)$ after it is factored.
    \item \textbf{Solution A (Full Expansion):}
    \begin{itemize}
        \item Step 1: Expand each trinomial product using the distributive property.
        \item Step 2: Add the expanded polynomials together and combine like terms.
        \item Step 3: Extract the constant term from the resulting polynomial.
    \end{itemize}
    \item \textbf{Solution B (Constant Term Shortcut):}
    \begin{itemize}
        \item Step 1: Determine the constant term of each product by multiplying the constant terms of the individual polynomials.
        \item Step 2: Add the constant terms from each product to find the constant term of the entire expression.
        \item Step 3: Verify that the constant term remains unchanged when the polynomial is factored.
    \end{itemize}
\end{itemize}

\textbf{Case Study 2: Summaries for Figure~\ref{fig:pca_example_2}}
\begin{itemize}
    \item \textbf{Question:} Determine the largest real value of $a$ such that the equation
    $$ax = x^3 + 1$$
    has a real solution.
    \item \textbf{Solution A (Calculus Approach):}
    \begin{itemize}
        \item Step 1: Rewrite the equation to express $ a $ as a function of $ x $, $ a = \frac{x^3 + 1}{x} $.
        \item Step 2: Find the critical points of the function $ f(x) = \frac{x^3 + 1}{x} $ by taking its derivative and setting it to zero.
        \item Step 3: Evaluate the function at the critical point to find the value of $ a $ where the equation has a double root, ensuring the largest $ a $ for which the equation has a real solution.
    \end{itemize}
    \item \textbf{Solution B (Geometric Interpretation):}
    \begin{itemize}
        \item Step 1: Interpret the equation as the intersection of a line $y = ax$ and a curve $y = x^3 + 1$.
        \item Step 2: Set the derivative of the cubic function equal to the slope of the line to find the point of tangency.
        \item Step 3: Solve the system of equations to find the largest real value of $ a $ corresponding to the tangency condition.
    \end{itemize}
\end{itemize}

\textbf{Analysis:}
The two case studies in Figure~\ref{fig:pca_plots} illustrate a consistent pattern where our RPD metric succeeds and the baseline fails. In both examples, the solution pairs employ fundamentally different strategies. The baseline Raw Embedding Distance assigns very low scores (0.015 and 0.016) that correspond to mediocre percentiles (44-52\%). This indicates the method is unable to reliably distinguish these solutions from the vast majority of superficially similar pairs. In stark contrast, our RPD metric assigns high scores (0.259 and 0.274) that fall into high percentiles (87-90\%), correctly identifying the significant strategic divergence. The PCA visualizations visually corroborate this finding: the well-separated step embeddings in both (a) and (b) confirm that the solutions follow distinct reasoning paths, a fact that only RPD consistently captures.

\clearpage

\section{Experiment Implementation Details}
\label{app:experiment_details}

\subsection{RPD Metric Evaluation (Details for Sec.~\ref{sec:rpd_metric_eval})}
\label{app:rpd_metric_eval_details}

In this section, we provide the implementation details for the RPD metric evaluation, including the prompts used for the LLM-based baseline and the evaluation judge.

\subsubsection{Prompt for the LLM-Selection Baseline}
\label{app:exp1_llm_select}
To create the ``LLM Selection'' baseline, we prompt the Qwen3-14B model to identify the most diverse pair of solutions from all available candidates for a given problem. The prompt is designed to encourage a focus on strategic differences rather than superficial text variations.

\begin{tcolorbox}[
    breakable,
    colback=gray!10!white,   % Light gray background
    colframe=gray!75!black,  % Dark gray frame
    title=Prompt for Selecting the Most Methodologically Diverse Solution Pair,
    fonttitle=\bfseries,
    arc=2mm,
    boxrule=1pt
]

You are a master mathematician and an expert in pedagogical analysis. Your task is to analyze multiple proposed solutions for a given problem and select a single pair of answers that represents the maximum possible methodological diversity. If no such pair exists, you must indicate this.

Your goal is to identify \textbf{one pair of answers} that represents a significant difference in a core step or sub-methodology. If all solutions follow a fundamentally similar strategy, your answer will be to select \textbf{``No"}.

\hrulefill
\subsubsection*{1. Your Analysis Framework \& Core Criteria}
\hrulefill

Your primary task is to act as a discerning analyst. You must distinguish between minor procedural choices and significant differences in core steps. Assume that most solutions might share a high-level strategy; your goal is to find answers that execute core steps in a meaningfully different way.

\paragraph{Defining Methodological Difference (Your Core Criteria):}

\paragraph{\textbf{What IS NOT a Significant Difference (Methodologically Similar):}}
\begin{itemize}
    \item \textbf{Order of Calculation:} Calculating value A then B, versus B then A, before combining them in the same way.
    \item \textbf{Algebraic Equivalence:} Using the form $(a+b)^2$ versus $a^2 + 2ab + b^2$.
    \item \textbf{Variable Naming or Notation:} Using $n$ vs $x$.
    \item \textbf{Choice of Standard Procedural Equivalents:} One summary describes solving a system of equations using \textbf{substitution}, while the other uses \textbf{elimination}. These are considered standard, interchangeable procedures within the same overall algebraic approach.
    \item \textbf{Rigorous Proof vs. Heuristic Assumption:} If the overall strategy is the same, simply proving a result versus assuming it does not constitute a diverse approach. Both are still following the same high-level logical path.
\end{itemize}

\paragraph{\textbf{What IS a Significant Difference (Methodologically Diverse):}}
\begin{itemize}
    \item This difference represents a \textbf{completely distinct, independent, high-level strategic choice} that fundamentally alters the entire problem-solving path from beginning to end.
    \item \textbf{Example 1 (Different Overall Framework):} One solution to a geometry problem uses \textbf{coordinate geometry}, another uses \textbf{synthetic geometry}, and a third uses \textbf{vector analysis}.
    \item \textbf{Example 2 (Completely Different Logical Path):} To solve a counting problem, one answer uses \textbf{direct casework}, another uses \textbf{complementary counting}, and a third uses a \textbf{recurrence relation}.
    \item \textbf{Example 3 (Change in Analytical Tool):} A solution to an optimization problem uses \textbf{calculus}, a second uses \textbf{inequalities} (like AM-GM), and a third uses \textbf{linear programming}.
\end{itemize}

\hrulefill
\subsubsection*{2. Content to Analyze}
\hrulefill

\textbf{Problem:}\\
\texttt{\{question\}}

\vspace{1em}
\textbf{Proposed Solutions (Summarized by Logical Steps):}\\
\texttt{\{summaries\_text\}}

\hrulefill
\subsubsection*{3. Final Instructions \& Output Requirement}
\hrulefill

\textbf{Your Task:}\\
Based on the final criteria review, analyze the solutions and make one of two possible determinations:
\begin{enumerate}
    \item Identify the single pair of answers with the maximum methodological diversity.
    \item Conclude that no pair meets the criteria for significant diversity, meaning all solutions follow a fundamentally similar approach.
\end{enumerate}

\paragraph{\textbf{Step 1: Brief Comparative Analysis}}
\begin{itemize}
    \item \textbf{If you find a diverse pair:} Write a single, brief paragraph. Do not summarize each solution individually. Instead, group the solutions by common methodology and justify your selection of the most diverse pair. For example: ``Solution A uses direct casework, while Solution B uses complementary counting. This represents the most significant methodological difference."
    \item \textbf{If you do NOT find a diverse pair:} Write a single, brief paragraph explaining why. State that all solutions follow a similar core strategy and briefly describe that common approach. For example: ``All solutions utilize a system of linear equations to solve for the variables. While they use different methods like substitution or elimination, this does not represent a significant strategic divergence. Therefore, no pair is methodologically diverse."
\end{itemize}

\paragraph{\textbf{Step 2: Final JSON Output}}
Immediately after your brief analysis paragraph, provide your final answer in a strict JSON format within a special block.
\begin{itemize}
    \item \textbf{If a diverse pair is found:} The JSON should be a list containing the single selected answer ID pair.
    \item \textbf{If no diverse pair is found:} The JSON should contain the string ``No" within the list structure to maintain format consistency.
\end{itemize}

\vspace{0.5em}
\textbf{Example of Final Output Structure (Diverse Pair):}\\
{[Your brief analysis justifying the choice...]}
\begin{verbatim}
//boxed_json{{[[id_A, id_B]]}}
\end{verbatim}

\vspace{0.5em}
\textbf{Example of Final Output Structure (No Diverse Pair):}\\
{[Your brief analysis explaining the lack of diversity...]}
\begin{verbatim}
//boxed_json{{[["No"]]}}
\end{verbatim}

Begin Analysis and Provide Output:

\end{tcolorbox}
\clearpage
\subsubsection{The LLM Evaluation Judge}
\label{app:exp1_llm_judge}
To automate the calculation of the ``success rate," a LLM Judge (Qwen3-14B) is used to provide a final verdict on the diversity of a solution pair selected by a given method (e.g., RPD, Raw Emb., etc.). This section details the prompt used to guide the judge and the study conducted to validate its alignment with human judgment.

\paragraph{Judge Prompt.} The judge is provided with the problem statement and a single pair of solutions. Its task is to assess whether the two solutions employed genuinely different problem-solving strategies. The prompt explicitly instructs the judge to ignore minor differences in wording or calculation and focus on the core reasoning approach.

\begin{tcolorbox}[
    breakable,
    colback=gray!10!white,   % Light gray background
    colframe=gray!75!black,  % Dark gray frame
    title=Prompt for Methodological Similarity Rating,
    fonttitle=\bfseries,
    arc=2mm,
    boxrule=1pt
]

You are an expert Answer Analysis Assistant, specializing in understanding and comparing the logic and methodology behind problem-solving. Your task is to receive a question, two full answers with their summaries, and rate them strictly based on the similarity of their \textbf{methodology}.

\textbf{Note:} Based on your prior analysis, you should assume that all proposed solutions for this problem follow a similar high-level strategy. Your task is to find and rate the \textbf{methodological diversity within this shared high-level strategy}.

\hrulefill
\subsubsection*{Rating Criteria}
\hrulefill

Your task is to determine if the two answers are \textbf{Methodologically Similar} or \textbf{Methodologically Diverse} based on the criteria below, and assign a corresponding rating.

\begin{itemize}
    \item \textbf{Rating 1 (Methodologically Similar):} The two answers are considered similar if the differences are superficial. The following are \textbf{NOT} considered significant methodological differences:
    \begin{itemize}
        \item \textbf{Order of Calculation:} Calculating value A then B, versus B then A, before combining them in the same way.
        \item \textbf{Algebraic Equivalence:} Using the form \texttt{(a+b)\textasciicircum2} versus \texttt{a\textasciicircum2 + 2ab + b\textasciicircum2}.
        \item \textbf{Variable Naming or Notation:} Using $n$ vs $x$.
        \item \textbf{Choice of Standard Procedural Equivalents:} One summary describes solving a system of equations using \textbf{substitution}, while the other uses \textbf{elimination}. These are considered standard, interchangeable procedures within the same overall algebraic approach.
        \item \textbf{Rigorous Proof vs. Heuristic Assumption:} If the overall strategy is the same, simply proving a result versus assuming it does not constitute a diverse approach. Both are still following the same high-level logical path.
    \end{itemize}

    \item \textbf{Rating 2 (Methodologically Diverse):} The two answers are considered diverse if the difference represents a \textbf{completely distinct, independent, high-level strategic choice} that fundamentally alters the entire problem-solving path from beginning to end.
    \begin{itemize}
        \item \textbf{Example 1 (Different Overall Framework):} One solution to a geometry problem uses \textbf{coordinate geometry}, another uses \textbf{synthetic geometry}, and a third uses \textbf{vector analysis}.
        \item \textbf{Example 2 (Completely Different Logical Path):} To solve a counting problem, one answer uses \textbf{direct casework}, another uses \textbf{complementary counting}, and a third uses a \textbf{recurrence relation}.
        \item \textbf{Example 3 (Change in Analytical Tool):} A solution to an optimization problem uses \textbf{calculus}, a second uses \textbf{inequalities} (like AM-GM), and a third uses \textbf{linear programming}.
    \end{itemize}
\end{itemize}

\hrulefill
\subsubsection*{Output Requirement}
\hrulefill

First, provide a detailed analysis explaining the methodological similarities and differences based on the criteria above. After your analysis is complete, provide the final rating on a new line in the format \texttt{//boxed\{\{rating\_number\}\}}. \textbf{DO NOT ONLY GIVE OUT YOUR RATE!}

\hrulefill

\textbf{Begin Analysis:}

\vspace{1em}
\textbf{[Question]:}\\
\texttt{\{question\}}

\vspace{1em}
\textbf{[Answer A]:}\\
\texttt{\{answer\_a\}}

\vspace{1em}
\textbf{[Answer A summary]:}\\
\texttt{\{summary\_a\}}

\vspace{1em}
\textbf{[Answer B]:}\\
\texttt{\{answer\_b\}}

\vspace{1em}
\textbf{[Answer B summary]:}\\
\texttt{\{summary\_b\}}

\end{tcolorbox}

\paragraph{Validation.}
To ensure the reliability of the LLM Judge used as our primary evaluation criterion in Sec.~\ref{sec:rpd_metric_eval}, we conduct an alignment study with human annotations.

\begin{wraptable}{r}{0.5\textwidth}
\vspace{-15pt} 
\centering
\caption{Confusion matrix of LLM Judge verdicts against human annotations on 100 solution pairs.}
\label{tab:judge_alignment}
\small 
\begin{tabular}{@{}llcc@{}}
\toprule
          &            & \multicolumn{2}{c}{\textbf{LLM Judge Verdict}} \\ 
          \cmidrule(lr){3-4}
          &            & Diverse         & Same            \\ \midrule
\textbf{Human} & \textbf{Diverse} & 41 (TP)         & 9 (FN)          \\
\textbf{Label} & \textbf{Same}    & 13 (FP)         & 37 (TN)         \\ \bottomrule
\end{tabular}
\end{wraptable}

To validate the judge, we first construct a dedicated test set. Human annotators select 100 pairs of solutions from our candidate pool, creating a balanced ground-truth dataset composed of 50 pairs with semantically \textit{diverse} reasoning paths and 50 pairs with the \textit{same} underlying reasoning path.

The LLM Judge is then tasked with making a binary diversity judgment on each of these 100 pairs. The results are presented in the confusion matrix in Table~\ref{tab:judge_alignment}. Overall, the LLM Judge achieves an accuracy of 78\%, demonstrating a strong alignment with human judgment and performing significantly better than a random baseline (50\%). We observe that the judge is quite effective at identifying truly diverse pairs (Recall 82\%), though it is slightly prone to false positives (classifying similar paths as diverse). This level of agreement validates our use of the LLM Judge as a reliable automated proxy for evaluating reasoning diversity in our main experiment.

\subsection{Details for Multi-Solution Fine-Tuning (Sec.~\ref{sec:main_exp})}
\label{app:baselines}

This section provides detailed implementation procedures for the main fine-tuning experiment, focusing on how the baseline training sets were constructed. Each method aims to select 100 problems and 3 solutions per problem, but they differ in their core selection strategy.

\subsubsection{Random Selection Baseline}

The \textbf{Random 1P3S} baseline was constructed through a naive sampling process. We first randomly selected 100 problems from our 1,600-problem candidate pool without replacement. For each of these 100 problems, we then randomly selected 3 of its available solutions to form the training data. This method serves as a fundamental baseline to measure the benefits of any systematic diversity-driven selection.

\subsubsection{LLM Selection Baseline}

This baseline leverages the powerful Qwen3-14B model to simulate an expert's judgment in a two-stage curation process. First, the LLM performs a binary classification to identify whether a problem's solutions are methodologically diverse. We then selected 100 problems that were positively classified as containing diverse solution methods. Second, for these selected problems, the LLM is prompted again to choose the set of 3 solutions that are maximally distinct from each other. The specific prompts for each stage are provided below.

\begin{tcolorbox}[
    breakable,
    colback=gray!10!white,   % Light gray background
    colframe=gray!75!black,  % Dark gray frame
    title=Prompt for Problem Diversity Classification,
    fonttitle=\bfseries,
    arc=2mm,
    boxrule=1pt
]

You are a master mathematician and an expert in pedagogical analysis. Your task is to classify a problem based on the methodological diversity of its proposed solutions.

Your goal is to perform a binary classification:
\begin{itemize}
    \item \textbf{Class 2 (Diverse):} If the provided solution summaries showcase more than one distinct core methodology.
    \item \textbf{Class 1 (Not Diverse):} If all solutions use the same core methodology, or if the differences are only superficial (e.g., a different order of calculation, or using standard procedural equivalents like substitution vs. elimination).
\end{itemize}

\vspace{2mm}
\hrule
\vspace{1mm}
\subsubsection*{1. Your Analysis Framework \& Core Criteria}

Your primary task is to act as a discerning analyst. You must distinguish between minor procedural choices and significant differences in core steps. Assume that most solutions might share a high-level strategy; your goal is to find answers that execute core steps in a meaningfully different way.

\paragraph{Defining Methodological Difference (Your Core Criteria):}

\paragraph{\textbf{What IS NOT a Significant Difference (Methodologically Similar):}}
\begin{itemize}
    \item \textbf{Order of Calculation:} Calculating value A then B, versus B then A, before combining them in the same way.
    \item \textbf{Algebraic Equivalence:} Using the form $(a+b)^2$ versus $a^2 + 2ab + b^2$.
    \item \textbf{Variable Naming or Notation:} Using $n$ vs $x$.
    \item \textbf{Choice of Standard Procedural Equivalents:} One summary describes solving a system of equations using \textbf{substitution}, while the other uses \textbf{elimination}. These are considered standard, interchangeable procedures within the same overall algebraic approach.
    \item \textbf{Rigorous Proof vs. Heuristic Assumption:} If the overall strategy is the same, simply proving a result versus assuming it does not constitute a diverse approach. Both are still following the same high-level logical path.
\end{itemize}

\paragraph{\textbf{What IS a Significant Difference (Methodologically Diverse):}}
\begin{itemize}
    \item This difference represents a \textbf{completely distinct, independent, high-level strategic choice} that fundamentally alters the entire problem-solving path from beginning to end.
    \item \textbf{Example 1 (Different Overall Framework):} One solution to a geometry problem uses \textbf{coordinate geometry}, another uses \textbf{synthetic geometry}, and a third uses \textbf{vector analysis}.
    \item \textbf{Example 2 (Completely Different Logical Path):} To solve a counting problem, one answer uses \textbf{direct casework}, another uses \textbf{complementary counting}, and a third uses a \textbf{recurrence relation}.
    \item \textbf{Example 3 (Change in Analytical Tool):} A solution to an optimization problem uses \textbf{calculus}, a second uses \textbf{inequalities} (like AM-GM), and a third uses \textbf{linear programming}.
\end{itemize}

\vspace{2mm}
\hrule
\vspace{1mm}
\subsubsection*{2. Content to Analyze}

\textbf{Problem:}\\
\{question\}

\vspace{3mm}
\textbf{Proposed Solutions (Summarized by Logical Steps):}\\
\{summaries\_text\}

\vspace{2mm}
\hrule
\vspace{1mm}
\subsubsection*{3. Output Requirement}

Based on the final criteria review, classify the diversity of the solutions.

\paragraph{Output Requirement:}
Immediately after your classification, provide your final answer in a strict JSON format within a special block. The JSON should be a single integer, either $1$ or $2$. Do not provide any other text.

\vspace{2mm}
\textbf{Example of Final Output Structure for a Diverse problem:}\\
\texttt{//boxed\{\{2\}\}}

\vspace{2mm}
\textbf{Example of Final Output Structure for a Not Diverse problem:}\\
\texttt{//boxed\{\{1\}\}}

\vspace{2mm}
\textbf{Begin Analysis and Provide Output:}

\end{tcolorbox}

\vspace{5mm}

\begin{tcolorbox}[
    breakable,
    colback=gray!10!white,   % Light gray background
    colframe=gray!75!black,  % Dark gray frame
    title=Prompt for Diverse Solution Selection,
    fonttitle=\bfseries,
    arc=2mm,
    boxrule=1pt
]
You are a master mathematician and an expert in pedagogical analysis. Your task is to analyze multiple proposed solutions for a given problem and select a set of \{num\_to\_select\} answers that, as a set, represents the maximum possible methodological diversity.

Your goal is to identify a single set of \{num\_to\_select\} answers where each chosen answer has a significant methodological difference from every other answer in the set. Think of it as finding a set of three solutions that are all mutually distinct in their core approach.

\vspace{2mm}
\hrule
\vspace{1mm}
\subsubsection*{1. Your Analysis Framework \& Core Criteria}

Your primary task is to act as a discerning analyst. You must distinguish between minor procedural choices and significant differences in core steps. Assume that most solutions might share a high-level strategy; your goal is to find answers that execute core steps in a meaningfully different way.

\paragraph{Defining Methodological Difference (Your Core Criteria):}

\paragraph{\textbf{What IS NOT a Significant Difference (Methodologically Similar):}}
\begin{itemize}
    \item \textbf{Order of Calculation:} Calculating value A then B, versus B then A, before combining them in the same way.
    \item \textbf{Algebraic Equivalence:} Using the form $(a+b)^2$ versus $a^2 + 2ab + b^2$.
    \item \textbf{Variable Naming or Notation:} Using $n$ vs $x$.
    \item \textbf{Choice of Standard Procedural Equivalents:} One summary describes solving a system of equations using \textbf{substitution}, while the other uses \textbf{elimination}. These are considered standard, interchangeable procedures within the same overall algebraic approach.
    \item \textbf{Rigorous Proof vs. Heuristic Assumption:} If the overall strategy is the same, simply proving a result versus assuming it does not constitute a diverse approach. Both are still following the same high-level logical path.
\end{itemize}

\paragraph{\textbf{What IS a Significant Difference (Methodologically Diverse):}}
\begin{itemize}
    \item This difference represents a \textbf{completely distinct, independent, high-level strategic choice} that fundamentally alters the entire problem-solving path from beginning to end.
    \item \textbf{Example 1 (Different Overall Framework):} One solution to a geometry problem uses \textbf{coordinate geometry}, another uses \textbf{synthetic geometry}, and a third uses \textbf{vector analysis}.
    \item \textbf{Example 2 (Completely Different Logical Path):} To solve a counting problem, one answer uses \textbf{direct casework}, another uses \textbf{complementary counting}, and a third uses a \textbf{recurrence relation}.
    \item \textbf{Example 3 (Change in Analytical Tool):} A solution to an optimization problem uses \textbf{calculus}, a second uses \textbf{inequalities} (like AM-GM), and a third uses \textbf{linear programming}.
\end{itemize}

\vspace{2mm}
\hrule
\vspace{1mm}
\subsubsection*{2. Content to Analyze}

\textbf{Problem:}\\
\{question\}

\vspace{3mm}
\textbf{Proposed Solutions (Summarized by Logical Steps):}\\
\{summaries\_text\}

\vspace{2mm}
\hrule
\vspace{1mm}
\subsubsection*{3. Final Instructions \& Output Requirement}

Your Task: Based on the final criteria review, analyze the solutions.

\paragraph{Step 1: Brief Comparative Analysis}
First, write a single, brief paragraph for your analysis. Do not summarize each solution individually. Instead, group the solutions by common methodology and justify your selection of the set of \{num\_to\_select\} most diverse answers. For example, Solutions A and C use direct casework, while Solution B uses complementary counting, and Solution D uses a geometric approach. The most diverse set is [A, B, D] as it captures these three distinct methods.

\paragraph{Step 2: Final JSON Output}
Immediately after your brief analysis paragraph, provide your final answer in a strict JSON format within a special block. The JSON should be a list containing the \{num\_to\_select\} selected answer IDs.

\vspace{2mm}
\textbf{Example of Final Output Structure:}\\
{[Your brief analysis...]}

\texttt{//boxed\_json\{\{[id\_A, id\_B, id\_C]\}\}}

\vspace{2mm}
\textbf{Begin Analysis and Provide Output:}
\end{tcolorbox}

\subsubsection{Embedding-Based Baseline}

To rigorously evaluate the effectiveness of our RPD metric, we compare it against two baseline distance metrics. For a fair comparison, all training datasets, both for our method and the baselines, are constructed using the identical \textbf{two-stage data curation framework} detailed previously. This framework consists of \textbf{Stage 1: Problem Selection} (Algorithm~\ref{alg:problem_selection_compact}) and \textbf{Stage 2: Greedy Solution Selection} (Algorithm~\ref{alg:greedy_maxmin_compact}).

The sole difference between our method and the baselines is the specific pairwise distance function, $\mathcal{D}(S_i, S_j)$, that is plugged into this framework. The baseline metrics are defined below.

\paragraph{Raw Solution Cosine Distance ($D_{\text{raw}}$)}
This baseline metric computes the cosine distance between the embedding vectors of the complete solution texts. For all embedding tasks, we use the Qwen3-Embedding-8B model. Let $\mathcal{M}_{\text{embed}}$ be this model.
$$D_{\text{raw}}(S_i, S_j) = 1 - \frac{\mathcal{M}_{\text{embed}}(S_i) \cdot \mathcal{M}_{\text{embed}}(S_j)}{\|\mathcal{M}_{\text{embed}}(S_i)\| \|\mathcal{M}_{\text{embed}}(S_j)\|}$$

\paragraph{Summary Cosine Distance ($D_{\text{summary}}$)}
This baseline first concatenates the step-level summaries for a solution to form a single composite summary text. The diversity is then computed as the cosine distance between the embeddings of these composite summaries.
$$D_{\text{summary}}(S_i, S_j) = 1 - \frac{\mathcal{M}_{\text{embed}}(\text{Summary}_{\text{comp}}(S_i)) \cdot \mathcal{M}_{\text{embed}}(\text{Summary}_{\text{comp}}(S_j))}{\|\mathcal{M}_{\text{embed}}(\text{Summary}_{\text{comp}}(S_i))\| \|\mathcal{M}_{\text{embed}}(\text{Summary}_{\text{comp}}(S_j))\|}$$

Based on the framework detailed previously, we generate three distinct training datasets:
\begin{itemize}
    \item \textbf{Ours (RPD)}: Constructed by applying the two-stage framework with our proposed RPD metric ($D_{\text{RPD}}$).
    \item \textbf{Raw Emb.}: Constructed using the same framework but with the $D_{\text{raw}}$ metric.
    \item \textbf{Summary Emb.}: Constructed using the same framework but with the $D_{\text{summary}}$ metric.
\end{itemize}

\clearpage

\section{Experiment Results}
\label{app:full_results}

This appendix presents the complete experimental results for both models. The tables are structured to clearly distinguish between the pre-trained baseline model, fine-tuning with a one-problem-one-solution (1P1S) paradigm, and fine-tuning with a one-problem-three-solution (1P3S) paradigm.

\subsection{Complete Results for Qwen3-4B-Base Model}
\label{app:full_results_qwen3_4b}
The following tables present the comprehensive performance of the Qwen3-4B-Base model on the AIME24 and Olympiad Benchmarks, which complements the MATH500 Level 5 results from the main paper. As shown in Table~\ref{tab:appendix_full_qwen3_aime}, our \textbf{RPD} method demonstrates a significant performance improvement by adopting the \textit{one problem, multiple solutions} paradigm. It elevates the \texttt{pass@16} score to \textbf{35.83\%} on AIME24, surpassing the standard 1P1S baseline (Random 1P1S) by an impressive \textbf{4.99} percentage points. Furthermore, our RPD-guided curation strategy also proves its superiority over other 1P3S methods, with its \texttt{pass@16} score outperforming the next-best baseline (Random 1P3S) by \textbf{2.50} percentage points on the same benchmark. This pattern holds for the Olympiad Bench (Table~\ref{tab:appendix_full_qwen3_olympiad}), where our method achieves a leading \texttt{pass@16} score of \textbf{68.11\%}, which is \textbf{1.56} percentage points higher than the 1P1S baseline and \textbf{0.75} percentage points higher than the best alternative 1P3S method. These results provide strong evidence for the effectiveness of our approach in both paradigm and data curation strategy.

\begin{table}[h!]
\centering
\small
\setlength{\tabcolsep}{4pt}
\caption{Full comparison on the \textbf{AIME24} benchmark using the \textbf{Qwen3-4B-Base} model.}
\label{tab:appendix_full_qwen3_aime}
\begin{tabular}{llccccc}
\toprule
\textbf{Paradigm} & \textbf{Method} & \textbf{pass@1 (\%)} & \textbf{pass@2 (\%)} & \textbf{pass@4 (\%)} & \textbf{pass@8 (\%)} & \textbf{pass@16 (\%)} \\
\midrule
Pre-trained & Base & 8.34 & 13.33 & 16.67 & 21.67 & 27.50 \\
\midrule
1P1S & Random 1P1S & \textbf{14.17} & 18.33 & 23.33 & 26.67 & 30.84 \\
\midrule
1P16S & Unfiltered & 6.67 & 12.50 & 20.83 & 25.83 & 29.17 \\
\midrule
\multirow{5}{*}{1P3S} & Random 1P3S & 9.17 & 12.50 & 19.17 & 28.34 & 33.33 \\
& Raw Emb. & 12.50 & 16.67 & 20.00 & 25.84 & 33.33 \\
& Summary Emb. & 10.00 & 12.50 & 17.50 & 25.00 & 29.17 \\
& LLM Selection & 10.83 & 15.84 & 20.83 & 25.83 & 30.83 \\
& \textbf{Ours (RPD)} & \textbf{14.17} & \textbf{19.17} & \textbf{25.83} & \textbf{30.00} & \textbf{35.83} \\
\bottomrule
\end{tabular}
\end{table}

\begin{table}[h!]
\centering
\small
\setlength{\tabcolsep}{4pt}
\caption{Full comparison on the \textbf{Olympiad Bench} using the \textbf{Qwen3-4B-Base} model.}
\label{tab:appendix_full_qwen3_olympiad}
\begin{tabular}{llccccc}
\toprule
\textbf{Paradigm} & \textbf{Method} & \textbf{pass@1 (\%)} & \textbf{pass@2 (\%)} & \textbf{pass@4 (\%)} & \textbf{pass@8 (\%)} & \textbf{pass@16 (\%)} \\
\midrule
Pre-trained & Base & 39.54 & 47.11 & 53.56 & 61.13 & 65.95 \\
\midrule
1P1S & Random 1P1S & \textbf{42.43} & 49.18 & 55.49 & 61.43 & 66.55 \\
\midrule
1P16S & Unfiltered & 39.47 & 49.11 & 56.61 & 62.99 & 68.03 \\
\midrule
\multirow{5}{*}{1P3S} & Random 1P3S & 40.13 & 50.15 & 56.75 & 62.61 & 67.36 \\
& Raw Emb. & 39.91 & 47.48 & 56.38 & 61.42 & 66.62 \\
& Summary Emb. & 40.88 & 49.78 & 57.05 & 62.69 & 66.92 \\
& LLM Selection & 39.62 & 48.30 & 56.60 & 62.83 & 67.06 \\
& \textbf{Ours (RPD)} & 41.92 & \textbf{51.19} & \textbf{57.50} & \textbf{63.06} & \textbf{68.11} \\
\bottomrule
\end{tabular}
\end{table}
%------------------------------------------------------------------------------------%

\subsection{Complete Results for Qwen2.5-3B Model}
\label{app:full_results_qwen2}
To demonstrate the robustness and generalizability of our findings, we also fine-tuned the Qwen2.5-3B model. Specifically, we employed supervised fine-tuning using 4-bit QLoRA (rank=16, alpha=32), training the model for \textbf{15 epochs} in BF16 precision. We utilized the AdamW optimizer with a cosine learning rate scheduler, setting the peak learning rate to $4 \times 10^{-5}$. We then evaluated its performance across the same three benchmarks (Tables~\ref{tab:appendix_full_qwen2.5_aime}, \ref{tab:appendix_full_qwen2.5_math}, and \ref{tab:appendix_full_qwen2.5_olympiad}). 

The results consistently reaffirm our core hypothesis. For instance, on the AIME24 benchmark (Table~\ref{tab:appendix_full_qwen2.5_aime}), our \textbf{RPD} method's advantage is particularly pronounced when evaluating with a larger sample set. Focusing on the key \texttt{pass@16} metric, our approach achieves a score of \textbf{22.50\%}. This represents a substantial 5.00 percentage point improvement over the 1P1S baseline and demonstrates a clear advantage over other multi-solution strategies, outperforming the next-best 1P3S methods by 0.83 percentage points. The outperformance on AIME24 exemplifies a consistent trend also observed on the MATH500 and Olympiad benchmarks, which solidifies the conclusion that our RPD-guided data curation is a general and effective technique for enhancing Test-Time Scaling.

\begin{table}[h!]
\centering
\small
\setlength{\tabcolsep}{4pt}
\caption{Full comparison on the \textbf{AIME24} benchmark using the \textbf{Qwen2.5-3B} model.}
\label{tab:appendix_full_qwen2.5_aime}
\begin{tabular}{llccccc}
\toprule
\textbf{Paradigm} & \textbf{Method} & \textbf{pass@1 (\%)} & \textbf{pass@2 (\%)} & \textbf{pass@4 (\%)} & \textbf{pass@8 (\%)} & \textbf{pass@16 (\%)} \\
\midrule
Pre-trained & Base & 4.17 & 4.17 & 10.00 & 16.67 & 16.67 \\
\midrule
1P1S & Random 1P1S & 4.17 & 8.34 & 10.00 & 13.33 & 17.50 \\
\midrule
1P16S & Unfiltered & 3.34 & 6.67 & 11.67 & 13.33 & 21.67 \\
\midrule
\multirow{5}{*}{1P3S} & Random 1P3S & 6.67 & 8.33 & 14.17 & 18.33 & 20.00 \\
& Raw Emb. & 5.84 & 8.34 & 14.17 & 18.33 & 20.83 \\
& Summary Emb. & 3.33 & 6.67 & 13.33 & 18.33 & 21.67 \\
& LLM Selection & 2.50 & 5.00 & 13.33 & 16.67 & 21.67 \\
& \textbf{Ours (RPD)} & \textbf{7.50} & \textbf{10.00} & \textbf{15.00} & \textbf{20.00} & \textbf{22.50} \\
\bottomrule
\end{tabular}
\end{table}
%-----------------------------------------------------------------------------%

\begin{table}[h!]
\centering
\small
\setlength{\tabcolsep}{4pt}
\caption{Full comparison on the \textbf{MATH500 Level 5} benchmark using the \textbf{Qwen2.5-3B} model.}
\label{tab:appendix_full_qwen2.5_math}
\begin{tabular}{llccccc}
\toprule
\textbf{Paradigm} & \textbf{Method} & \textbf{pass@1 (\%)} & \textbf{pass@2 (\%)} & \textbf{pass@4 (\%)} & \textbf{pass@8 (\%)} & \textbf{pass@16 (\%)} \\
\midrule
Pre-trained & Base & 23.70 & 32.65 & 43.84 & 55.60 & 63.62 \\
\midrule
1P1S & Random 1P1S & 29.11 & 41.05 & 51.31 & 60.45 & 67.73 \\
\midrule
1P16S & Unfiltered & 29.29 & 39.74 & 51.31 & 59.33 & 66.98 \\
\midrule
\multirow{5}{*}{1P3S} & Random 1P3S & \textbf{31.72} & \textbf{42.91} & 50.94 & 60.82 & 68.28 \\
& Raw Emb. & 28.92 & 40.86 & 51.31 & 60.08 & 69.22 \\
& Summary Emb. & 27.05 & 38.06 & 51.12 & 60.26 & 67.35 \\
& LLM Selection & 27.61 & 37.87 & 49.82 & 60.26 & 67.91 \\
& \textbf{Ours (RPD)} & 28.55 & 40.30 & \textbf{51.49} & \textbf{61.20} & \textbf{69.97} \\
\bottomrule
\end{tabular}
\end{table}
%---------------------------------------------------------------------------------%

\begin{table}[h!]
\centering
\small
\setlength{\tabcolsep}{4pt}
\caption{Full comparison on the \textbf{Olympiad Bench} using the \textbf{Qwen2.5-3B} model.}
\label{tab:appendix_full_qwen2.5_olympiad}
\begin{tabular}{llccccc}
\toprule
\textbf{Paradigm} & \textbf{Method} & \textbf{pass@1 (\%)} & \textbf{pass@2 (\%)} & \textbf{pass@4 (\%)} & \textbf{pass@8 (\%)} & \textbf{pass@16 (\%)} \\
\midrule
Pre-trained & Base & 21.81 & 30.27 & 37.54 & 45.55 & 51.93 \\
\midrule
1P1S & Random 1P1S & 19.14 & 27.45 & 35.68 & 45.48 & 52.89 \\
\midrule
% 1P16S & Unfiltered & 20.33 & 29.75 & 38.95 & \textbf{47.26} & 53.93 \\
% \midrule
\multirow{5}{*}{1P3S} & Random 1P3S & 22.33 & 30.79 & 39.10 & 47.11 & 53.93 \\
& Raw Emb. & 22.03 & 30.05 & 39.10 & 46.52 & 52.90 \\
& Summary Emb. & \textbf{22.85} & \textbf{31.34} & 38.95 & 46.63 & 53.82 \\
& LLM Selection & 21.96 & 30.79 & \textbf{39.25} & 47.11 & 53.94 \\
& \textbf{Ours (RPD)} & 20.40 & 30.19 & 39.10 & 47.18 & \textbf{54.16} \\
\bottomrule
\end{tabular}
\end{table}

\clearpage
\subsection{Complete Results for Llama-3.1-8B-Instruct Model}
\label{app:full_results_llama3}
To further validate our approach on a different architecture, we extended our experiments to the \textbf{Llama-3.1-8B-Instruct} model. We maintained a similar supervised fine-tuning setup using 4-bit QLoRA (rank=16, alpha=32) and BF16 precision. For this model, we adjusted the training duration to \textbf{24 epochs} and set the peak learning rate to $5 \times 10^{-5}$ with the AdamW optimizer and a cosine scheduler. The model was then evaluated on the same three mathematical reasoning benchmarks (Tables~\ref{tab:appendix_full_llama3_combined}).

The results from Llama-3.1-8B-Instruct further corroborate the effectiveness of our RPD-guided data selection. The benefits of our approach are especially pronounced on the \textbf{MATH500 Level 5} benchmark. Our method consistently outperforms the 1P1S baseline across all sampling levels, achieving the most significant gain at \texttt{pass@8} with a score of \textbf{47.76\%} — a \textbf{3.35 percentage point} improvement over the baseline. This strong performance trend extends to the other challenging benchmarks, with our model also demonstrating clear advantages over the 1P1S baseline on AIME24 and Olympiad Bench. These findings strongly indicate that our RPD-based data curation is a generalizable and effective strategy for enhancing Test-Time Scaling, independent of the base model architecture.

\begin{table}[h!]
\centering
\small
\setlength{\tabcolsep}{3pt}
\caption{Performance comparison of our 1PNS approach (RPD) against the 1P1S baseline (Random) using Llama-3.1-8B-Instruct across three mathematical reasoning benchmarks.}
\label{tab:appendix_full_llama3_combined}
\begin{tabular}{llccccc}
\toprule
\textbf{Benchmark} & \textbf{Method} & \textbf{pass@1 (\%)} & \textbf{pass@2} & \textbf{pass@4} & \textbf{pass@8} & \textbf{pass@16} \\
\midrule
\multirow{2}{*}{AIME24} 
& Random (1P1S) & 3.75  & 4.58  & 8.75  & 12.50 & 16.67 \\
& RPD (1P3S)   & \textbf{6.67}  & \textbf{8.75}  & \textbf{11.25} & \textbf{15.00} & \textbf{18.75} \\
\midrule
\multirow{2}{*}{MATH500 Level 5} 
& Random (1P1S) & 17.35 & 26.31 & 35.45 & 44.41 & 52.62 \\
& RPD (1P3S)   & \textbf{20.52} & \textbf{28.36} & \textbf{37.69} & \textbf{47.76} & \textbf{55.23} \\
\midrule
\multirow{2}{*}{Olympiad Bench} 
& Random (1P1S) & 13.99 & 21.29 & 26.86 & 33.46 & 39.62 \\
& RPD (1P3S)   & \textbf{14.54} & \textbf{21.59} & \textbf{27.45} & \textbf{33.83} & \textbf{40.58} \\
\bottomrule
\end{tabular}
\end{table}

\clearpage
\subsection{Experimental Details for RL Fine-Tuning}
\label{app:rl_details}

We performed an additional phase of Reinforcement Learning (RL) fine-tuning on our SFT checkpoints. The training was conducted for 3 epochs using the Group Relative Policy Optimization (GRPO; \citealp{deepseekai2025deepseekr1incentivizingreasoningcapability}) algorithm implemented in the veRL framework. Key hyperparameters and configuration details for this stage are summarized in Table~\ref{tab:rl_hyperparams}.

\begin{table}[h!]
\centering
\small
\caption{Hyperparameters for Reinforcement Learning Fine-Tuning.}
\label{tab:rl_hyperparams}
\begin{tabular}{ll}
\toprule
\textbf{Hyperparameter} & \textbf{Value} \\
\midrule
\multicolumn{2}{l}{\textit{\textbf{Algorithm \& Framework}}} \\
Framework & veRL \\
Algorithm & Group Relative Policy Optimization (GRPO) \\
Dataset & AIME (1983--2023) \\
Training Epochs & 3 \\
KL Coefficient ($\lambda_{KL}$) & 0.001 \\
KL in Reward & False \\
\midrule
\multicolumn{2}{l}{\textit{\textbf{Actor Model \& Optimization}}} \\
Actor Learning Rate & $2 \times 10^{-5}$ \\
LoRA Rank & 32 \\
LoRA Alpha & 16 \\
LoRA Target Modules & All linear layers \\
\midrule
\multicolumn{2}{l}{\textit{\textbf{Data \& Generation Configuration}}} \\
Max Prompt Length & 3800 tokens \\
Max Response Length & 10000 tokens \\
Max Model Length (vLLM) & 13800 tokens \\
Rollout Engine & vLLM \\
Rollout Samples ($n$) & 8 \\
Rollout Temperature & 0.7 \\
\bottomrule
\end{tabular}
\end{table}

\clearpage

\subsection{Ablation Study: Performance at a Larger Scale (3000 Samples)}
\label{app:scale_up_ablation}

To further assess the scalability of our 1PNS paradigm, we extended our training data to a total of 3,000 samples. In this experiment, we compared our diversity-driven approach (1,000 questions, 3 solutions each) against a traditional 1P1S baseline (3,000 unique questions, 1 solution each). Models in this ablation were trained for 1 epoch.

The results on Qwen3-4B-Base, summarized in Table~\ref{tab:scale_up}, demonstrate that our RPD-curated 1P3S approach consistently outperforms the 1P1S baseline across all mathematical reasoning benchmarks. Notably, our method achieves superior performance in almost all metrics, particularly in higher $k$ values (pass@4 to pass@16), confirming that the benefits of multi-solution fine-tuning remain robust and effective as the data scale increases.

\begin{table}[h!]
\centering
\small
\setlength{\tabcolsep}{3pt}
\caption{Performance comparison of our 1P3S approach against the 1P1S baseline on Qwen3-4B-Base at a larger scale (3,000 samples) across benchmarks.}
\label{tab:scale_up}
\begin{tabular}{llccccc}
\toprule
\textbf{Benchmark} & \textbf{Method} & \textbf{pass@1 (\%)} & \textbf{pass@2} & \textbf{pass@4} & \textbf{pass@8} & \textbf{pass@16} \\
\midrule
\multirow{2}{*}{AIME24} & Random (1P1S) & \textbf{10.00} & \textbf{16.67} & 20.83 & 25.83 & 32.50 \\
 & Ours (RPD 1P3S) & \textbf{10.00} & \textbf{16.67} & \textbf{25.00} & \textbf{29.17} & \textbf{35.83} \\
% \midrule
% \multirow{2}{*}{AIME25} & Random (1P1S) & 9.17 & 14.17 & 19.17 & 23.34 & 32.50 \\
%  & Ours (RPD 1P3S) & \textbf{10.00} & \textbf{15.83} & \textbf{24.17} & \textbf{30.84} & \textbf{36.67} \\
\midrule
\multirow{2}{*}{MATH500 Level 5} & Random (1P1S) & 50.56 & 58.96 & 66.23 & 73.13 & 77.61 \\
 & Ours (RPD 1P3S) & \textbf{52.05} & \textbf{60.64} & \textbf{70.34} & \textbf{76.49} & \textbf{80.04} \\
\midrule
\multirow{2}{*}{Olympiad Bench} & Random (1P1S) & 40.06 & 47.77 & 55.64 & 61.57 & 66.91 \\
 & Ours (RPD 1P3S) & \textbf{40.50} & \textbf{49.26} & \textbf{58.01} & \textbf{63.06} & \textbf{68.84} \\
% \midrule
% \multirow{2}{*}{AMC23} & Random (1P1S) & 41.25 & \textbf{51.25} & 65.00 & 76.25 & 83.75 \\
%  & Ours (RPD 1P3S) & \textbf{42.50} & 50.00 & \textbf{68.75} & \textbf{78.75} & \textbf{87.50} \\
\bottomrule
\end{tabular}
\end{table}

\newpage
\subsection{Computational Overhead Analysis}
\label{app:compute_overhead}

To evaluate the scalability and efficiency of our data curation pipeline, we profiled the computational cost on 2x H20 GPUs. Table~\ref{tab:compute_cost} details the time consumption for each stage of the process.

The total time to curate a single solution using our RPD pipeline is approximately \textbf{1.68 seconds}. This comprises 1.64s for LLM Summarization (using Qwen3-14B on the OpenThought dataset) and 0.04s for Embedding (using Qwen3-Embedding-8B). The computational cost for the final pairwise distance calculation is negligible (approximately 5.6ms per problem). We argue that this overhead is highly acceptable for two key reasons:

\begin{enumerate}
    \item \textbf{Efficiency Relative to Generation:} The curation cost (1.68s) is significantly lower than the inference time required to generate a single Long CoT solution (approx. 4.04s on a 4B model). This ratio becomes even more favorable when larger models (e.g., 32B+) are used for data generation, making the curation overhead a minor fraction of the total pipeline.
    \item \textbf{One-Time Cost for Extensive Reuse:} The RPD curation is strictly a \textit{one-time construction cost}. Once the high-quality dataset is built, it can be reused extensively by the community for repeated training runs. Compared to the cumulative compute resources required for these downstream training processes, the initial single-pass curation cost is highly acceptable.
\end{enumerate}

\begin{table}[h!]
\centering
\small
\caption{Computational cost breakdown per solution. Our RPD pipeline incurs a modest overhead compared to the raw baseline.}
\label{tab:compute_cost}
\begin{tabular}{llc}
\toprule
\textbf{Pipeline} & \textbf{Component} & \textbf{Time / Solution} \\
\midrule
\multirow{3}{*}{Ours (RPD)} & LLM Summarization (Qwen3-14B) & 1.64s \\
 & Embedding (Qwen3-Embedding-8B) & 0.04s \\
 \cmidrule(l){2-3}
 & \textbf{Total Cost} & \textbf{1.68s} \\
\midrule
Baseline & Raw Embedding (Qwen3-Embedding-8B) & 0.91s \\
\bottomrule
\end{tabular}
\end{table}

\newpage
\subsection{Ablation Study: Robustness of RPD to Summarizer Model Choice}
\label{app:summarizer_ablation}

To evaluate the stability of our RPD metric calculation, we conducted an ablation study to assess whether our pipeline relies on a specific large-scale model for the "Reasoning Step Extraction" phase (Section~\ref{sec:method}). In this experiment, we replaced the original summarizer (Qwen3-14B) with a significantly smaller model, \textbf{Qwen2.5-7B-Instruct}, while keeping the downstream fine-tuning model (Qwen3-4B-Base) and all other training settings unchanged.

The results, presented in Table~\ref{tab:summarizer_robustness}, demonstrate that the performance of our method using the 7B summarizer is highly comparable to, and in some cases exceeds, that of the 14B summarizer. Crucially, both RPD-based configurations consistently outperform the Random 1P1S baseline across varying $k$ values. This confirms that our RPD pipeline is robust and not overly dependent on the capability of the summarization model. We attribute this stability to our detailed structured prompt (Appendix~\ref{app:summarization_prompt}), which enables even smaller instruction-tuned models to extract reasoning patterns reliably.

\begin{table}[h!]
\centering
\small
\setlength{\tabcolsep}{3pt}
\caption{Robustness analysis of the RPD pipeline. Performance comparison using different models for the summarization step (Qwen3-14B vs. Qwen2.5-7B-Instruct) against the baseline.}
\label{tab:summarizer_robustness}
\begin{tabular}{llccccc}
\toprule
\textbf{Benchmark} & \textbf{Method (Summarizer)} & \textbf{pass@1 (\%)} & \textbf{pass@2} & \textbf{pass@4} & \textbf{pass@8} & \textbf{pass@16} \\
\midrule
\multirow{3}{*}{AIME24} & Random (1P1S) & \textbf{14.17} & 18.33 & 23.33 & 26.67 & 30.84 \\
 & Ours (Qwen3-14B Sum.) & \textbf{14.17} & 19.17 & 25.83 & \textbf{30.00} & 35.83 \\
 & Ours (Qwen2.5-7B Sum.) & 13.33 & \textbf{22.50} & \textbf{27.50} & 29.17 & \textbf{36.67} \\
\midrule
\multirow{3}{*}{MATH500 L5} & Random (1P1S) & 49.26 & 60.64 & 66.98 & 72.20 & 77.43 \\
 & Ours (Qwen3-14B Sum.) & 52.61 & 61.57 & \textbf{71.64} & \textbf{75.94} & 79.29 \\
 & Ours (Qwen2.5-7B Sum.) & \textbf{53.73} & \textbf{63.81} & 68.66 & 75.75 & \textbf{79.67} \\
\midrule
\multirow{3}{*}{Olympiad} & Random (1P1S) & \textbf{42.43} & 49.18 & 55.49 & 61.43 & 66.55 \\
 & Ours (Qwen3-14B Sum.) & 41.92 & \textbf{51.19} & \textbf{57.50} & \textbf{63.06} & \textbf{68.11} \\
 & Ours (Qwen2.5-7B Sum.) & 38.58 & 48.07 & 55.34 & 62.61 & 67.80 \\
\bottomrule
\end{tabular}
\end{table}

\newpage
\subsection{Generalization Beyond Math: Code Generation}
\label{app:code_generalization}

To evaluate the universality of the 1PNS paradigm, we extended our RPD pipeline beyond the mathematical domain to code generation.

\paragraph{Experimental Setup.} We curated a dataset consisting of 300 code training samples from OpenThought3. We constructed two training sets: (1) a diversity-driven set using our RPD method (1P3S), and (2) a random baseline set (1P1S). We combined these code samples with the corresponding math datasets and fine-tuned the \textbf{Qwen3-4B-Base} model for 10 epochs to demonstrate the applicability of our paradigm to the code generation domain.

\paragraph{Results.} We evaluated the models on our original math benchmarks as well as two code benchmarks: \textbf{Live Code Bench} and \textbf{HumanEval}. The results, presented in Table~\ref{tab:code_generalization}, show that our method consistently outperforms both the Base Model and the Random 1P1S baseline across all evaluated datasets in both domains. Notably, on HumanEval, our method achieves a significant improvement in pass@16 (87.20\% vs. 80.73\%) compared to the random baseline. This confirms that the RPD pipeline and the 1PNS paradigm are generalizable principles effectively enhancing reasoning diversity in code generation tasks.

\begin{table}[h!]
\centering
\small
\setlength{\tabcolsep}{3pt}
\caption{Performance comparison on both Math and Code benchmarks. The models were fine-tuned on a mixed dataset (Math + Code). Our method demonstrates universal improvement across both domains.}
\label{tab:code_generalization}
\begin{tabular}{llccccc}
\toprule
\textbf{Benchmark} & \textbf{Method} & \textbf{pass@1 (\%)} & \textbf{pass@2} & \textbf{pass@4} & \textbf{pass@8} & \textbf{pass@16} \\
\midrule
\multicolumn{7}{c}{\textit{\textbf{Math Benchmarks}}} \\
\midrule
\multirow{3}{*}{AIME24} & Base Model & 8.34 & 13.33 & 16.67 & 21.67 & 27.50 \\
 & Random 1P1S (Math + Code) & 10.84 & 15.00 & 19.17 & 23.33 & 29.17 \\
 & \textbf{Ours (Math + Code)} & \textbf{12.50} & \textbf{16.67} & \textbf{22.50} & \textbf{26.67} & \textbf{31.67} \\
\midrule
\multirow{3}{*}{MATH500 L5} & Base Model & 46.08 & 56.90 & 64.37 & 71.27 & 75.00 \\
 & Random 1P1S (Math + Code) & 45.71 & 57.46 & 64.56 & 69.03 & 75.93 \\
 & \textbf{Ours (Math + Code)} & \textbf{48.13} & \textbf{57.84} & \textbf{67.91} & \textbf{72.57} & \textbf{77.24} \\
\midrule
\multirow{3}{*}{Olympiad Bench} & Base Model & 39.54 & 47.11 & 53.56 & 61.13 & 65.95 \\
 & Random 1P1S (Math + Code) & 39.84 & 47.92 & 53.94 & 60.61 & 66.18 \\
 & \textbf{Ours (Math + Code)} & \textbf{39.99} & \textbf{49.04} & \textbf{55.20} & \textbf{61.72} & \textbf{67.29} \\
\midrule
\multicolumn{7}{c}{\textit{\textbf{Code Benchmarks}}} \\
\midrule
\multirow{3}{*}{Live Code Bench} & Base Model & 13.46 & 22.46 & 30.05 & 35.35 & 38.86 \\
 & Random 1P1S (Math + Code) & 14.22 & 21.99 & 30.05 & 36.30 & 40.61 \\
 & \textbf{Ours (Math + Code)} & \textbf{17.35} & \textbf{25.88} & \textbf{32.04} & \textbf{38.10} & \textbf{42.56} \\
\midrule
\multirow{3}{*}{HumanEval} & Base Model & 2.86 & 5.36 & 9.56 & 15.93 & 25.00 \\
 & Random 1P1S (Math + Code) & 18.64 & 33.54 & 47.26 & 66.36 & 80.73 \\
 & \textbf{Ours (Math + Code)} & \textbf{23.36} & \textbf{38.74} & \textbf{57.73} & \textbf{75.44} & \textbf{87.20} \\
\bottomrule
\end{tabular}
\end{table}

\end{CJK*}
\end{document}